%% file: root.tex
\documentclass[journal]{IEEEtran}
\IEEEoverridecommandlockouts                              
\usepackage[english]{babel}
\usepackage{ifpdf}
\usepackage{url}
\usepackage{hyperref}
\usepackage{color}
\usepackage{pgf, tikz, pgfplots}
\usetikzlibrary{shapes, arrows, automata}
\usetikzlibrary{calc,hobby,decorations}
\usepackage[cmex10]{amsmath}
\usepackage{amsfonts, amssymb}
\usepackage{mathrsfs}
\usepackage[ruled,vlined]{algorithm2e} 
\usepackage{amsmath} 
\usepackage{xcolor}
\usepackage{enumerate}
\usepackage{multirow}
\usepackage{rotating}
\usepackage{subcaption}
\usepackage{booktabs}
\input{mySections.tex}

\input{mySymbol.sty}
\input{colors}

\def\SO3{\mathrm{SO(3)}}

\newtheorem{remark}{\hspace{0pt}\bf Remark}

\title{\LARGE \bf
        Differentiable Environment-Trajectory Co-Optimization for\\
        Safe Multi-Agent Navigation
}
\author{ 
}

\usepackage[backend=biber, style=ieee]{biblatex}
\pgfplotsset{compat=1.18}

\begin{document}
\author{Zhan Gao$^{\dagger, *}$, ~Gabriele Fadini$^{*}$, ~Stelian Coros, and Amanda Prorok 
	\thanks{
    $^{\dagger}$ Corresponding Author, $^{*}$ these Authors contributed equally. Z. Gao and A. Prorok are with the Department of Computer Science and Technology, University of Cambridge, UK. G. Fadini is with the Zurich University of Applied Sciences, ZHAW, Winterthur, Swizerland. S. Coros is with the Computational Robotics Lab, Department of Computer Science, ETH Zurich, Switzerland. This work is supported by ERC Project 949940 (gAIa).}}
\maketitle
\thispagestyle{empty}
\pagestyle{empty}
\begin{abstract}
    The environment plays a critical role in multi-agent navigation by imposing spatial constraints, rules, and limitations that agents must navigate around. Traditional approaches treat the environment as fixed, without exploring its impact on agents' performance. This work considers environment configurations as decision variables, alongside agent actions, to jointly achieve safe navigation. We formulate a bi-level problem, where the lower-level sub-problem optimizes agent trajectories that minimize navigation cost and the upper-level sub-problem optimizes environment configurations that maximize navigation safety. We develop a differentiable optimization method that iteratively solves the lower-level sub-problem with interior point methods and the upper-level sub-problem with gradient ascent. A key challenge lies in analytically coupling these two levels. We address this by leveraging KKT conditions and the Implicit Function Theorem to compute gradients of agent trajectories w.r.t. environment parameters, enabling differentiation throughout the bi-level structure. Moreover, we propose a novel metric that quantifies navigation safety as a criterion for the upper-level environment optimization, and prove its validity through measure theory. Our experiments validate the effectiveness of the proposed framework in a variety of safety-critical navigation scenarios, inspired from warehouse logistics to urban transportation. The results demonstrate that optimized environments provide navigation guidance, improving both agents' safety and efficiency.
\end{abstract}


\section{Introduction}

Multi-agent systems consist of multiple interactive agents that act within a shared environment, providing an effective framework for addressing spatially distributed tasks \cite{arai2002advances, ismail2018survey, gao2023online}. Ensuring safe and efficient navigation of multi-agent systems represents a fundamental challenge with broad applicability, encompassing domains such as autonomous vehicles, robotic systems, crowd simulation, and warehouse automation \cite{de2006formation, van2008reciprocal, desaraju2012decentralized, sartoretti2019primal, gao2024provably}. Multi-agent navigation requires agents to move from initial positions to designated goals while minimizing traveled distance or energy consumption, and simultaneously avoiding collisions with obstacle regions and other agents. Navigation performance is typically evaluated from two aspects: safety and efficiency. The former concerns the agents’ ability to avoid collisions with obstacles, hazards, and other agents, while the latter pertains to their capability of moving towards goals in an optimal manner. 

\begin{figure}[t!]%
	\centering
	\begin{subfigure}{0.45\columnwidth}
            \centering
		\includegraphics[width=\textwidth]{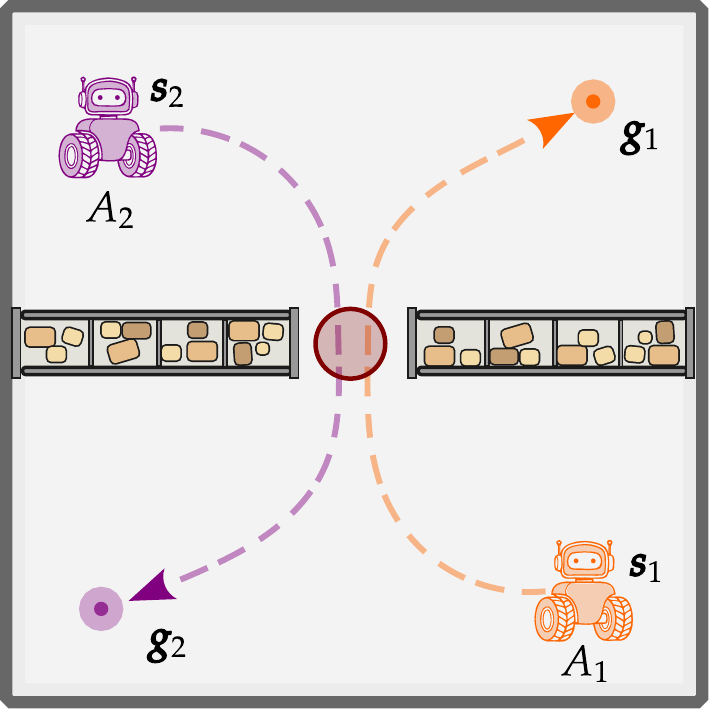}%
		\caption{Poorly-designed environment}%
		\label{subfig2a}%
	\end{subfigure}\hfill\hfill
	\begin{subfigure}{0.45\columnwidth}
            \centering
		\includegraphics[width=\textwidth]{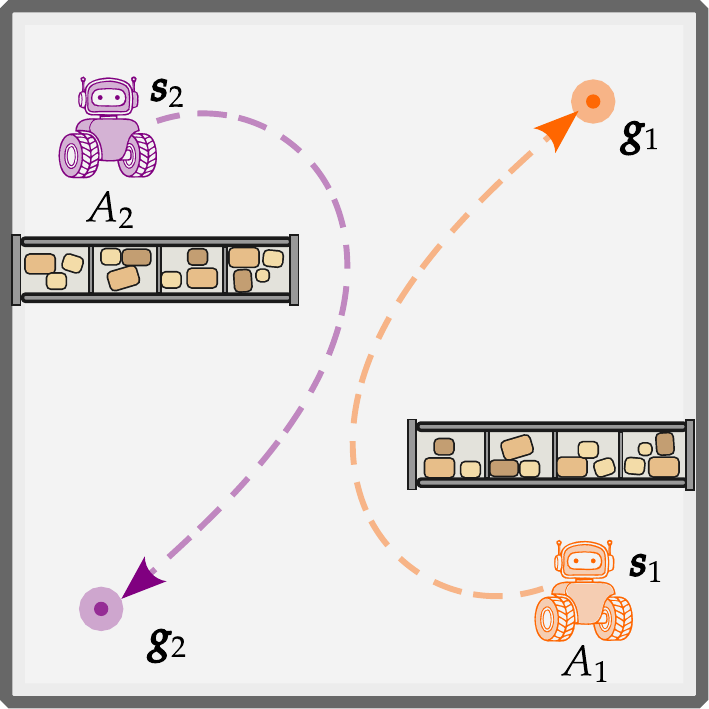}%
		\caption{Well-designed environment}%
		\label{subfig2b}%
	\end{subfigure}
	\caption{Environment configurations can impact the safety and efficiency of agent trajectories. Moving obstacle regions further apart in the environment of 
    Fig. \ref{subfig2b} allows to generate agent trajectories that are less likely to collide and exhibit improved path efficiency, compared to those in the environment of Fig. \ref{subfig2a}.}\label{fig:environment-design-relationship}\vspace{-4mm}
\end{figure}

Existing literature primarily focuses on developing effective navigation algorithms, while these algorithms consider the agents’ environment as fixed spatial constraints that must be circumvented. A crucial, yet frequently overlooked, aspect lies in the interplay between environment configurations and agent behaviors, i.e., the layout of obstacle regions can profoundly influence the agents' ability to reach their goals safely and efficiently. In particular, spatial constraints of poorly-designed environments may result in irresolvable navigation outcomes, such as dead-locks, live-locks, and prioritization conflicts, even for state-of-the-art algorithms \cite{mani2010search, ruderman2019uncovering}. To handle such bottlenecks, spatial structures (e.g., intersections and roundabouts) and markings (e.g., lanes) are used to facilitate agent de-confliction. However, their designs are often rooted in legacy mobility paradigms and ignore agent–environment interactions as well as system-level optimization. 

Reconfigurable environments are emerging as a new trend, in which environment configurations can be adjusted intelligently to better support multi-agent systems \cite{wang2010new, bier2014robotic, custodio2020flexible}. For instance, shelf locations in warehouses can be adjusted using rack pulleys or sliding track systems, and traffic routes in smart cities can be redesigned through urban planning initiatives. This creates opportunities to reconfigure the spatial layout of the environment as an additional means to improve the performance of multi-agent navigation, which is particularly appealing when agents are expected to perform repetitive tasks in structured environments (e.g., warehouses, factories, urban transportation systems). It is common practice to hand-design environment configurations, even though this may be inefficient and sub-optimal, especially in the continuous domain encountered in practical applications \cite{vcap2015prioritized}. 

This work aims to consider environment configurations as decision variables, alongside agent trajectories, in a system-level optimization framework to jointly improve the performance of multi-agent navigation. Specifically, optimizing environment configurations and generating agent trajectories are deeply interdependent problems within multi-agent navigation. As showcased in Fig.~\ref{fig:environment-design-relationship}, spatial locations of obstacles can have a significant impact on agent trajectories, compromising their safety and efficiency. Conversely, agent trajectories can reveal how suitable an environment is for executing navigation tasks and guide the optimal tuning of environment configurations. Applications include warehouse logistics (e.g., determining optimal shelf positions for collision-free cargo transportation \cite{wurman2008coordinating}), city planning (e.g., designing road layouts for safe multi-vehicle driving \cite{drezner1997selecting}), search and rescue (e.g., clearing optimal passages for trapped victims to safely escape \cite{karma2015use}), and digital entertainment (e.g., creating optimal gaming scenes for smooth movement of non-player characters \cite{johnson2001computer}).

Towards this end, we propose an environment-trajectory co-optimization framework tailored for safe multi-agent navigation. The goal is to optimize the spatial layout of obstacle regions to facilitate the generation of collision-free agent trajectories. More in detail, our contributions are:

\smallskip
\noindent \emph{\textbf{(i)}} We define a bi-level environment-trajectory co-optimization problem, which considers both environment configurations and agent trajectories as decision variables. The lower level generates collision-free agent trajectories to improve efficiency, while the upper level builds upon the former and optimizes obstacle regions to enhance safety. This bi-level formulation preserves the optimality of the lower-level trajectory optimization, while accounting for this optimality during the upper-level environment optimization.  

\smallskip
\noindent \emph{\textbf{(ii)}} We develop a principled differentiable optimization method for the bi-level problem. It solves the lower-level trajectory optimization with interior point methods, computes gradients of the generated trajectories w.r.t. environment parameters with Karush–Kuhn–Tucker (KKT) conditions and the Implicit Function Theorem (IFT), and solves the upper-level environment optimization with gradient-based methods. This characterizes explicitly the relationship between the environment and agents by computing analytic gradients, which enables to connect the lower-level sub-problem with the upper-level sub-problem for differentiable bi-level optimization. 

\smallskip
\noindent \emph{\textbf{(iii)}} We propose a novel safety metric that quantifies an explicit safety level of the environment w.r.t. multi-agent navigation in a continuous manner. The metric transcends the binary definition of navigation safety in the literature and allows to distinguish environments with higher or lower collision risks. Moreover, it satisfies fundamental properties grounded in measure theory, which ensures its mathematical validity as a criterion for the upper-level environment optimization. 

\smallskip
\noindent \emph{\textbf{(iv)}} We validate our framework with extensive experiments across diverse scenarios of safe multi-agent navigation. The results corroborate theoretical findings and demonstrate substantial improvements in navigation safety, efficiency, and computational cost. This highlights the potential of our method to inform the practical design of traffic systems and smart cities, paving the way for safer and more reliable multi-agent systems.

\smallskip
\noindent \textbf{Related Works.} 
The concept of co-optimization in robotics originates from research on embodied cognitive science \cite{clark1998being}, which advocates designing intelligent systems in which control policies, physical structures, and environmental interactions are co-optimized to achieve task efficiency. The works in \cite{tatikonda2004control, tanaka2015sdp, tzoumas2020lqg} explore this concept to jointly design sensing strategies and motion controllers for achieving desired performance, while \cite{lipson2000automatic, cheney2018scalable} leverage evolutionary approaches to simultaneously optimize manufacturing, morphology, and control of embodied robots. More recent works use bi-level optimization to efficiently decouple the problem of agent design and control \cite{fadini_co-designing_2024, fadini_computational_2021, fadini_simulation_2022}, and exploit sensitivity analysis to recover meaningful optimization gradients \cite{de_vincenti_control-aware_2021, ha_computational_2018, ha_joint_2017, digumarti_concurrent_2014}. In the domain of multi-agent systems, \cite{gabel2008joint} presents a joint equilibrium policy search method that encourages agents to cooperate to reach states maximizing a global reward. The authors in \cite{zhang2016co} develop a multi-abstraction search approach to co-optimize agent placement with task assignment and scheduling, while \cite{jaleel2016decentralized, ali2018motion} consider joint optimization of mobility and communication, and design coordination strategies to minimize the total energy. However, these works cover only the design of robot bodies and controllers, leaving the problem of jointly optimizing agent behaviors and environment configurations rather unexplored.

Environment configurations have a significant impact on agent behaviors \cite{gao2023environment, gao2023constrained, gao2025co}. The works in \cite{jager2001decentralized, bennewitz2002finding} demonstrate that there exist congestion and dead-locks in undesirable environments, and develop methods that coordinate agents to avoid potential dead-locks. The authors in \cite{vcap2015complete, vcap2015prioritized} propose the concept of ``well-formed environment'' in which navigation tasks of agents can be carried out successfully without collisions, while \cite{wu2020multi} identifies the impact of the environment shape on agent trajectories and generates distinct path prospects for different agents to coordinate their motion. Gur et al. \cite{gur2021adversarial} study adversarial environments and develop resilient navigation algorithms in these environments. While these works acknowledge that the design of robotic agents cannot be isolated from the environment in which they operate, none of them consider environment configurations as decision variables, as well as agent trajectories, to improve performance in a system-level optimization framework.

More similar to our work, \cite{hauser2013minimum, hauser2014minimum} remove obstacle constraints from the environment to improve navigation performance, but focus on a single agent scenario. Bellusci et al. \cite{bellusci2020multi} extend the concept to multi-agent systems and search over all possible environment configurations to find the best solution for agents, but focus on discrete settings and consider only obstacle removal in experiments. The works in \cite{gao2023environment, gao2023constrained, gao2025co} consider the continuous domain and leverage reinforcement learning to optimize environment configurations for multi-agent navigation. However, these works do not characterize an explicit relationship among the environment, agents, and performance, but consider this relationship as a black box and use search-based or learning-based mechanisms to guide environment optimization in a model-free manner, which can be computationally expensive and sub-optimal. 

\section{Problem Formulation}

We consider an environment $\ccalE(\bbvartheta)$ with $M$ obstacle regions $\{\Delta_j(\bbvartheta)\}_{j=1}^M$ parametrized by variables $\bbvartheta$ and a multi-agent system $\ccalA$ with $N$ agents $\{A_i\}_{i=1}^N$. The agents follow a navigation strategy from starting positions $\bbS = [\bbs_1,\ldots,\bbs_N]$ towards goals $\bbG = [\bbg_1,\ldots,\bbg_N]$, while avoiding collisions with both obstacle regions and other agents. Specifically, let $\bbx_i^{(t)}, \bbu_i^{(t)}$ be the state and action of $A_i$ at time step $t$, $T$ the total number of time steps, $\bbx_i = \{\bbx_i^{(t)}\}_{t=0}^T$ and $\bbu_i = \{\bbu_i^{(t)}\}_{t=0}^T$ the sequences of states and actions across time steps, and $d t$ the duration of each time step -- see Fig.~\ref{fig:environment-design-relationship} for an illustration. We assume that agent dynamics take the general form of 
\begin{align}\label{eq:dynamics}
	\bbx_i^{(t)} = \Phi(\bbx^{(t-1)}_i, \bbu^{(t-1)}_i, dt),~\text{for}~i=1,\ldots,N,
\end{align}
where $\Phi(\cdot)$ is the state transition map from the current state $\bbx^{(t-1)}_i$ to the next state $\bbx_i^{(t)}$ under the control action $\bbu^{(t-1)}_i$.

The goal of this work is twofold: (i) generate agent trajectories $\bbx, \bbu = \{\bbx_i\}_{i=1}^N, \{\bbu_i\}_{i=1}^N$ that maximize navigation efficiency while ensuring agents' safety in the environment $\ccalE(\bbvartheta)$, and (ii) optimize the layout of obstacle regions $\{\Delta_j(\bbvartheta)\}_{j=1}^M$ in the environment $\ccalE(\bbvartheta)$ to further facilitate safe multi-agent navigation. That is, we seek to design the optimal environment $\ccalE(\bbvartheta^\star)$, in which the optimal trajectories $\{\bbx^\star, \bbu^\star\}$ are generated to jointly maximize navigation performance across sub-objectives of path efficiency, control effort, and safety. 

Environment parameters $\bbvartheta$ and agent trajectories $\{\bbx, \bbu\}$ are implicitly dependent, i.e. $\{\bbx, \bbu\}$ are generated under spatial constraints determined by $\bbvartheta$, while $\bbvartheta$ need to be optimized based on the performance of the generated $\{\bbx, \bbu\}$. This motivates to formulate a bi-level problem, where the upper-level environment optimization builds upon the lower-level trajectory optimization. In the following, we first introduce the two sub-problems and then formulate the bi-level problem. 

\subsection{Multi-Agent Trajectory Optimization}\label{subsec:TO}

Collision-free multi-agent navigation can be achieved via trajectory optimization, which minimizes destination distance and energy consumption subject to system dynamics and collision-avoidance constraints as
\begin{align}\label{eq:ECBF-ZCBF-QP}
	\min_{\substack{ \bbx \in \mathbb{X} \\ \bbu \in \mathbb{U}}} \quad &
	f(\mathbf{x}, \mathbf{u}) = \sum_{i,t} \left\| \bbg_i - \bbx_i^{(t)} \right\|_{\bbR_1}^2 + \sum_{i,t} \left\| \bbu_i^{(t)} \right\|_{\bbR_2}^2 \\
	\text{s.t.} \quad & \text{\small Agent dynamics}
	~\;\; \bbx_i^{(t)} = \Phi(\bbx_i^{(t-1)}, \bbu_i^{(t-1)}, dt) \nonumber \\
        & \text{\small Obstacle avoidance} 
	~\;\; g_{o,i}\big(\bbx_i^{(t)}, \bbvartheta\big) \le 0 \nonumber \\
	& \text{\small Agent avoidance}
	~\;\; g_a(\bbx_{i}^{(t)}, \bbx_{i'}^{(t)}) \leq 0 ~\for~ i \neq i' & \nonumber \\
        & \text{\small Initial conditions} 
	~~\;\; \bbx_{i}^{(0)} = \bbs_{i},& \nonumber
\end{align}
where $i, i' \in \{1, ..., N\}$ are the agent indices, $t \in \{1, ..., T\}$ the time-step index, $\bbR_1$ and $\bbR_2$ the weighting matrices for regularization, $\mathbb{U}$ and $\mathbb{X}$ the control and state feasible spaces, and $g_{o,i}(\bbx_i^{(t)}, \bbvartheta)$ and $g_a(\bbx_{i}^{(t)}, \bbx_{i'}^{(t)})$ the collision avoidance constraints w.r.t. obstacle regions and among agents. The objective $f(\mathbf{x}, \mathbf{u})$ minimizes the distances to goal states (first term) and the magnitudes of agent actions (second term), corresponding respectively to path efficiency and control effort, while the constraints ensure dynamics compliance, obstacle avoidance, inter-agent avoidance, and initial conditions. 

In problem \eqref{eq:ECBF-ZCBF-QP}, both agent and obstacle avoidance constraints take general forms, which may have different formulations depending on agent and obstacle specifications. In this work, we focus on simplified differentiable representations of agents and obstacles. For example, $g_a(\bbx_{i}^{(t)}, \bbx_{i'}^{(t)}) \le 0$ takes the form of a quadratic constraint if agents can be encircled by a convex-hull outer approximation. Similarly, $g_{o,i}\big(\bbx_i^{(t)}, \bbvartheta\big) \le 0$ is also quadratic if the obstacle can be approximated by a circular region. In the case of road boundaries, we can impose constraints in the form of perpendicular line constraints. For irregular obstacle regions, we may assume, without loss of generality, that they can either be encircled by an outer circular approximation to compute a conservative distance or obtained by the composition of simpler differentiable constraints.

\subsection{Environment Optimization}\label{subsec:EO}

The optimal trajectories $\{\bbx^\star(\bbvartheta), \bbu^\star(\bbvartheta)\}$ of problem \eqref{eq:ECBF-ZCBF-QP} depend on obstacle regions $\{\Delta_j(\bbvartheta)\}_{j=1}^M$
and thus, are functions of environment parameters $\bbvartheta$. This implies that a well-designed environment with appropriate obstacle regions has the potential to facilitate safe multi-agent navigation and ease trajectory optimization. In this context, we formulate the sub-problem of environment optimization as
\begin{align}\label{eq:EnvOpt}
	& ~\underset{\bbvartheta \in \mathbb{O}}{\text{max}}
	& &  F\big(\bbx^\star(\bbvartheta), \bbu^\star(\bbvartheta), \bbvartheta\big)\\
	& ~\text{s.t.}
	& & G\big(\bbx^\star(\bbvartheta), \bbu^\star(\bbvartheta), \bbvartheta\big) \leq 0, \nonumber 
\end{align}
where $F$ is a metric w.r.t. safe multi-agent navigation, which will be detailed in Section \ref{sec:Metric}, $\mathbb{O}$ is the feasible space of environment parameters $\bbvartheta$, and the constraints $G$ represent both: \emph{(i)} conditions involving agent states (e.g., obstacle regions do not overlap with starting and goal positions of agents) and \emph{(ii)} conditions about the design space itself (e.g. related to feasible locations of obstacles).
Problem \eqref{eq:EnvOpt} builds upon the optimal trajectories of problem \eqref{eq:ECBF-ZCBF-QP}, i.e., $F$ and $G$ are functions of $\{\bbx^\star(\bbvartheta), \bbu^\star(\bbvartheta)\}$. Such a relationship among agents' optimal trajectories $\bbx^\star(\bbvartheta)$, $\bbu^\star(\bbvartheta)$ and environment parameters $\bbvartheta$ is not generally expressible with any closed-form formulation, resulting in the need of a bi-level framework. 

\subsection{Bi-Level Optimization}

We combine the sub-problems in Sections \ref{subsec:TO} and \ref{subsec:EO} to propose a bi-level environment-trajectory co-optimization problem for safe multi-agent navigation as 
\begin{align}\label{eq:bilevel}
    & \texttt{Upper-level problem (environment)} \nonumber\\
    \max_{\bbvartheta \in \mathbb{O}} \quad &
    F(\bbx^\star(\bbvartheta), \bbu^\star(\bbvartheta), \bbvartheta) \\
    \text{s.t.} \quad & \text{Obstacle compliance} ~~
    G\big(\bbx^\star(\bbvartheta), \bbu^\star(\bbvartheta), \bbvartheta\big) \leq 0 \nonumber \\
    & \texttt{Lower-level problem (trajectories)} \nonumber \\
    &
    \bbx^\star(\bbvartheta), \bbu^\star(\bbvartheta) = \argmin_{\bbx \in \mathbb{X},\bbu \in \mathbb{U}} f(\bbx, \bbu) \nonumber \\
    & \text{s.t.} ~~ \text{Agent dynamics}~~ \bbx_i^{(t)} = \Phi(\bbx_i^{(t-1)}, \bbu_i^{(t-1)}, dt) \nonumber \\
    & \qquad \text{Obstacle avoidance} 
    ~~ g_{o,i}\big(\bbx_i^{(t)}, \bbvartheta\big) \le 0 \nonumber \\
    & \qquad \text{Agent avoidance}~~g_a(\bbx_{i}^{(t)}, \bbx_{i'}^{(t)}) \leq 0 ~\for~ i \neq i' \nonumber \\
    & \qquad \text{Initial conditions} ~~ \bbx_{i}^{(0)} = \bbs_{i}.& \nonumber
\end{align}
It co-optimizes environment parameters $\bbvartheta$, agent states $\bbx$, and agent actions $\bbu$, comprising the upper-level sub-problem of environment optimization \eqref{eq:EnvOpt} and the lower-level sub-problem of trajectory optimization \eqref{eq:ECBF-ZCBF-QP}. Specifically, $\bbx, \bbu$ are first computed via the lower-level sub-problem given $\bbvartheta$, which determines the upper-level sub-problem, and then $\bbvartheta$ is optimized based on the computed $\bbx, \bbu$. 

This bi-level formulation decomposes the joint optimization into two sub-problems, which decouples the space of decision variables and reduces the problem complexity. Moreover, it preserves the lower-level optimality of trajectory optimization and allows the upper-level sub-problem taking this optimality into account during environment optimization. However, solving the bi-level problem \eqref{eq:bilevel} faces three key challenges:

\smallskip
\noindent \emph{\textbf{(i)}} The two sub-problems are tightly intertwined, with updates in one propagating through and reshaping the other. This induces deep mutual dependencies that render the bi-level optimization problem complex and difficult to solve.

\smallskip
\noindent \emph{\textbf{(ii)}} The upper-level metric $F$ depends on the optimal trajectories $\{\bbx^\star(\bbvartheta), \bbu^\star(\bbvartheta)\}$, which implicitly depends on the environment parameters $\bbvartheta$. However, no closed-form solution exists for $\{\bbx^\star(\bbvartheta), \bbu^\star(\bbvartheta)\}$ in the lower-level trajectory optimization \eqref{eq:ECBF-ZCBF-QP}, rendering the upper-level environment optimization \eqref{eq:EnvOpt} analytically intractable and challenging to handle. 

\smallskip
\noindent \emph{\textbf{(iii)}} Both upper- and lower-level sub-problems may be non-convex, which complicates the bi-level optimization problem.

\section{Differentiable Optimization Methodology}\label{sec:Method}

In light of the aforementioned challenges, we develop a differentiable optimization method. It first solves the lower-level trajectory optimization with interior point methods, and then leverages Karush–Kuhn–Tucker (KKT) conditions and the Implicit Function Theorem (IFT) to compute gradients of agent states / actions w.r.t. environment parameters. These parametric sensitivities are then used to improve environment parameters in the upper-level environment optimization with gradient-based methods. Moreover, these gradients offer analytical insights into the relationship between agents and their surrounding environment, which is conventionally considered unknown in the literature. 

\subsection{Assembling KKT Conditions}\label{subsec:KKT}

From the bi-level formulation \eqref{eq:bilevel}, the metric of the upper-level environment optimization depends on the solution of the lower-level trajectory optimization, i.e., $F(\mathbf{x}^{\star}(\bbvartheta), \mathbf{u}^{\star}(\bbvartheta), \bbvartheta)$ is a function of $\{\mathbf{x}^{\star}(\bbvartheta), \mathbf{u}^{\star}(\bbvartheta)\}$. This indicates that solving the bi-level problem requires first characterizing the relationship between agent trajectories $\{\mathbf{x}^{\star}(\bbvartheta), \mathbf{u}^{\star}(\bbvartheta)\}$ and environment parameters $\bbvartheta$. We approach this by computing the gradients of $\{\mathbf{x}^{\star}(\bbvartheta), \mathbf{u}^{\star}(\bbvartheta)\}$ w.r.t. $\bbvartheta$ using KKT conditions and the IFT. 

Specifically, we start by formulating the Lagrangian of the lower-level trajectory optimization \eqref{eq:ECBF-ZCBF-QP} as 
\begin{align}\label{eq:lagrangian}
	&L(\bbx, \bbu, \bbvartheta, \bblambda, \bbbeta, \bbalpha, \bbxi) = f(\bbx, \bbu)\\
	& + \sum_{t=1}^{T} \sum_{i=1}^N {\bblambda_{i}^{(t)}}^\top \big(\bbx_{i}^{(t)} - \Phi(\bbx_i^{(t-1)}, \bbu_i^{(t-1)}, dt)\big) \nonumber\\
	& + \sum_{t=0}^T \!\sum_{i=1}^N \beta_i^{(t)} g_{o,i}(\bbx_i^{(t)}, \bbvartheta) \nonumber \\
    &+ \sum_{t=0}^{T}\!\sum_{i \ne i'} \alpha_{i,i'}^{(t)} g_a(\bbx_{i}^{(t)}, \bbx_{i'}^{(t)}) \!+\! \sum_{i=1}^N {\bbxi_{i}}^\top (\bbx_{i}^{(0)}- \bbs_i), \nonumber
\end{align}
where $\{\bblambda_{i}^{(t)}\}_{i,t}$, $\{\beta_i^{(t)}\}_{i,t}$, $\{\alpha_{i,i'}^{(t)}\}_{i \ne i', t}$, and $\{\bbxi_{i}^{(t)}\}_{i,t}$ are dual variables corresponding to the constraints of system dynamics, collision avoidance, and initial conditions in \eqref{eq:ECBF-ZCBF-QP}, and $\bblambda, \bbbeta, \bbalpha, \bbxi$ represent the concatenated dual variable vectors. In this context, we can formulate the KKT conditions as
\begin{align}
	&\label{eq:kkt1}\nabla_{\mathbf{x}} L(\mathbf{x},\! \mathbf{u},\! \bbvartheta,\! \bblambda,\! \bbbeta,\! \bbalpha,\! \bbxi) \!=\! \mathbf{0},~\!\nabla_{\mathbf{u}} L(\mathbf{x},\! \mathbf{u},\! \bbvartheta,\! \bblambda,\! \bbbeta,\! \bbalpha,\! \bbxi) \!=\! \mathbf{0},\\
	&\label{eq:kkt3}{\bblambda_{i}^{(t)}} \big(\bbx_{i}^{(t)} - \Phi(\bbx_i^{(t-1)}, \bbu_i^{(t-1)}, dt)\big)=\mathbf{0},~\forall~i~\text{and}~t,\\
    &\label{eq:kkt4} \beta_i^{(t)} g_{o,i}(\bbx_i^{(t)}, \bbvartheta)=0,~\forall~i~\text{and}~t,\\ 
	&\label{eq:kkt5}\alpha_{i,i'}^{(t)} g_a(\bbx_{i}^{(t)}, \bbx_{i'}^{(t)}) = 0,~\forall~i \ne i'~\text{and}~t,\\
	&\label{eq:kkt6}{\bbxi_{i}} (\bbx_{i}^{(0)}- \bbs_i)=\mathbf{0},~\forall~i.
\end{align}
These conditions \eqref{eq:kkt1}-\eqref{eq:kkt6} enable to leverage the IFT to compute the gradients of primal and dual variables $(\mathbf{x}, \mathbf{u}, \bblambda, \bbbeta, \bbalpha, \bbxi)$ w.r.t. environment parameters $\bbvartheta$. 

\subsection{Implicit Function Theorem}\label{subsec:Implicit}

When system states are linked to problem parameters through equality constraints, the IFT provides a tool that allows to compute gradients of system states w.r.t. problem parameters, which guarantees local differentiability and enables efficient computational approaches for differentiable optimization \cite{rudin_principles_2008}. Specifically, we consider agent trajectories, i.e., primal variables, and dual variables as functions of environment parameters $\bbx(\bbvartheta)$, $\bbu(\bbvartheta)$, $\bblambda(\bbvartheta)$, $\bbbeta(\bbvartheta)$, $\bbalpha(\bbvartheta)$, $\bbxi(\bbvartheta)$, and rewrite the KKT conditions as
\begin{align}
	&\label{eq:implicit_general}C\big(\bbvartheta, \mathbf{x}(\bbvartheta), \mathbf{u}(\bbvartheta), \bblambda(\bbvartheta), \bbbeta(\bbvartheta), \bbalpha(\bbvartheta), \bbxi(\bbvartheta)\big) = \mathbf{0},
\end{align}
where $C(\cdot)$ represents the concatenated KKT condition functions in 
\eqref{eq:kkt1}-\eqref{eq:kkt6}. We can compute the partial derivatives of \eqref{eq:implicit_general} w.r.t. environment parameters $\bbvartheta$ as
\begin{align}
	\mathbf{D}_{\bbvartheta} &= \frac{\partial C}{\partial \bbvartheta}.
\end{align}
The matrix dimension of $\mathbf{D}_{\bbvartheta}$ is the number of primal and dual variables times the number of environment parameters. The partial derivatives of \eqref{eq:implicit_general} w.r.t. primal and dual variables $\mathbf{x}, \mathbf{u}, \bblambda, \bbbeta, \bbalpha, \bbxi$ can be computed as 
\begin{align}
	\mathbf{D}_{\text{agent}} &= \frac{\partial C}{\partial (\mathbf{x}, \mathbf{u}, \bblambda, \bbbeta, \bbalpha, \bbxi)},
\end{align}
which is a square matrix. We define the partial derivatives of $\bbx(\bbvartheta)$, $\bbu(\bbvartheta)$, $\bblambda(\bbvartheta)$, $\bbbeta(\bbvartheta)$, $\bbalpha(\bbvartheta)$, and $\bbxi(\bbvartheta)$ w.r.t. $\bbvartheta$ as $\bbD_{\bbvartheta}^{\rm agt}$. By using the IFT with a first-order Taylor expansion, we get
\begin{align}
	\mathbf{D}_{\bbvartheta} + \mathbf{D}_{\text{agent}} \mathbf{D}_{\bbvartheta}^{\text{agent}} = \mathbf{0}.
\end{align}
Assuming $\mathbf{D}_{\text{agent}}$ is invertible, we obtain
\begin{align}\label{eq:implicitFunction}
	\mathbf{D}_{\bbvartheta}^{\text{agent}} = - \mathbf{D}_{\text{agent}}^{-1} \mathbf{D}_{\bbvartheta}.
\end{align}
This provides the gradients of agent trajectories w.r.t. environment parameters. These gradients characterize an explicit relationship among agent states $\bbx$, actions $\bbu$, and environment parameters $\bbvartheta$, a relationship that is conventionally unknown, and allow us to connect the lower-level trajectory optimization \eqref{eq:ECBF-ZCBF-QP} with the upper-level environment optimization \eqref{eq:EnvOpt}. 

\subsection{Differentiable Bi-Level Optimization}\label{subsec:DiffBilevel}

{\linespread{1.05}
\begin{algorithm}[t]
    \SetAlgoLined
    \KwIn{Initial environment parameters $\bbvartheta_0$, starting positions $\bbS$, goal positions $\bbG$, system dynamics $\Phi$, weighting matrices $\bbR_1$ and $\bbR_2$}
    \KwOut{Optimized environment parameters $\bbvartheta^\star$, agent trajectories $\bbx^\star$, and agent actions $\bbu^\star$}
    
    \For{$\kappa = 1, 2, \dots \ccalK$}{
        \tcp{Lower-level}
        Solve the sub-problem of trajectory optimization \eqref{eq:ECBF-ZCBF-QP} with interior point methods to obtain $\mathbf{x}^{\star}(\bbvartheta_\kappa)$ and $\mathbf{u}^{\star}(\bbvartheta_\kappa)$ \;

        Assemble the KKT conditions 
        \eqref{eq:lagrangian}\;
        
        Compute the gradients of $\mathbf{x}^{\star}(\bbvartheta_\kappa)$ and $\mathbf{u}^{\star}(\bbvartheta_\kappa)$ w.r.t. 
        $\bbvartheta_\kappa$, i.e., $\mathbf{D}_{\kappa, \bbvartheta}^{\text{agent}}$, via the IFT \eqref{eq:implicitFunction}\;
        
        \tcp{Upper-level}
        Compute the gradients of the metric function $F$ w.r.t. $\bbvartheta_\kappa$, i.e., $\nabla_{\bbvartheta} F (\mathbf{x}^{\star}(\bbvartheta_\kappa), \mathbf{u}^{\star}(\bbvartheta_\kappa), \bbvartheta_\kappa)$ \eqref{eq:derivative}\;
        
        Update environment parameters as $\bbvartheta_{\kappa+1} = \bbvartheta_\kappa + \Delta \alpha \nabla_{\bbvartheta} F (\mathbf{x}^{\star}(\bbvartheta_\kappa), \mathbf{u}^{\star}(\bbvartheta_\kappa), \bbvartheta_\kappa)$
        \;
    }
    \Return{$\bbvartheta^\star = \bbvartheta_\ccalK, \bbx^\star = \bbx^\star(\bbvartheta_\ccalK)$, and $\bbu^\star = \bbu^\star(\bbvartheta_\ccalK)$}
    \caption{Differentiable Bi-Level Optimization}
    \label{alg:DifferentiableMethod}
\end{algorithm}}

With the gradients $\mathbf{D}_{\bbvartheta}^{\text{agent}}$ obtained from Section \ref{subsec:Implicit}, we solve the bi-level problem with a differentiable optimization method, as explained in Algorithm \ref{alg:DifferentiableMethod}. Specifically, it consists of multiple iterations, where each iteration contains two phases: \emph{(I)} lower-level interior point optimization and \emph{(II)} upper-level gradient ascent. Details are presented as follows. 

\smallskip
\noindent \emph{Phase I.} At each iteration $\kappa$ with environment parameters $\bbvartheta_\kappa$, we formulate the lower-level sub-problem of trajectory optimization [cf. \eqref{eq:ECBF-ZCBF-QP}]. Given the non-linear nature of \eqref{eq:ECBF-ZCBF-QP}, we solve this sub-problem using interior point methods, which are standard to tackle non-convex optimization problems. This procedure yields an optimal solution $\{\mathbf{x}^{\star}(\bbvartheta_\kappa), \mathbf{u}^{\star}(\bbvartheta_\kappa)\}$ for multi-agent navigation in the environment $\ccalE(\bbvartheta_\kappa)$.\footnote{The solution may be locally optimal due to the non-convexity of the lower-level sub-problem.} 

With the optimal solution, we follow Sections \ref{subsec:KKT}-\ref{subsec:Implicit} to compute the gradients of the solution w.r.t. environment parameters $\mathbf{D}_{\kappa, \bbvartheta}^{\text{agent}}$ instantiated at $\{\mathbf{x}^{\star}(\bbvartheta_\kappa), \mathbf{u}^{\star}(\bbvartheta_\kappa)\}$ and $\bbvartheta_\kappa$. Then, we complete Phase I and step into Phase II. 

\smallskip
\noindent \emph{Phase II.} We formulate the upper-level sub-problem of environment optimization. Leveraging the gradients $\mathbf{D}_{\kappa, \bbvartheta}^{\text{agent}}$ from Phase I and the chain rule, we compute the gradients of the metric function $F$ w.r.t. environment parameters as
\begin{align}\label{eq:derivative}
	\nabla_{\bbvartheta} F (\mathbf{x}^{\star}(\bbvartheta_\kappa), \mathbf{u}^{\star}(\bbvartheta_\kappa), \bbvartheta_\kappa) 
	= \mathbf{D}^{F}_{\kappa, \text{agent}} \mathbf{D}^{\text{agent}}_{\kappa, \bbvartheta} + \mathbf{D}^{F}_{\kappa, \bbvartheta},
\end{align}
where $\mathbf{D}^{F}_{\kappa, \text{agent}}$ and $\mathbf{D}^{F}_{\kappa, \bbvartheta}$ are partial derivatives of $F$ w.r.t. agent trajectories and environment parameters, respectively, instantiated at $\{\bbx^\star(\bbvartheta_\kappa), \bbu^\star(\bbvartheta_\kappa)\}$ and $\bbvartheta_\kappa$. Here, $\mathbf{D}^{F}_{\kappa, \text{agent}}$ and $\mathbf{D}^{F}_{\kappa, \bbvartheta}$ can be computed directly because the explicit expression of $F$ is known by design choice, which will be introduced in Section \ref{sec:Metric}, while $\mathbf{D}^{\text{agent}}_{\kappa, \bbvartheta}$ is computed using KKT conditions and the IFT as detailed in Sections \ref{subsec:KKT}-\ref{subsec:Implicit}. This allows us to update $\bbvartheta_k$ with gradient ascent as
\begin{align}\label{eq:update}
	\bbvartheta_{\kappa+1} = \bbvartheta_{\kappa} + \Delta \alpha \nabla_{\bbvartheta} F(\mathbf{x}^{\star}(\bbvartheta_\kappa), \mathbf{u}^{\star}(\bbvartheta_\kappa), \bbvartheta_\kappa)
\end{align}
with $\Delta \alpha$ the step size. This completes iteration $\kappa$ and forwards the updated environment parameters $\bbvartheta_{\kappa+1}$ into iteration $\kappa+1$. 

\subsection{Motivation for a Safety Metric} 

The proposed differentiable bi-level optimization encodes navigation efficiency (i.e., path efficiency and energy consumption) in the objective $f$ of the lower-level trajectory optimization. The metric $F$ of the upper-level environment optimization \eqref{eq:EnvOpt} allows to account for performance assessment that explicitly depends on environment parameters $\bbvartheta$, beyond the objective $f$ of the lower-level trajectory optimization. While any valid metric can be applied in our framework, we focus on the safety of multi-agent navigation. 

In particular, we aim to optimize environment configurations to facilitate agent de-confliction and ease collision-free trajectory generation. While the lower-level trajectory optimization incorporates hard constraints for collision avoidance, it does not reveal an explicit safety level of the environment w.r.t. multi-agent navigation. For example, in Fig. \ref{fig:environment-design-relationship}, agents can navigate from $\bbs_1, \bbs_2$ to $\bbg_1, \bbg_2$ without collisions in both environments (a) and (b). However, in environment (a), agents are more likely to collide with either the obstacle region or each other, and it is more challenging to solve the lower-level sub-problem with trajectory optimization, compared with environment (b). This necessitates defining a new safety metric that explicitly measures collision risk of multi-agent navigation 
to serve as a criterion for the upper-level environment optimization. 

\section{Safety Metric Design}\label{sec:Metric}

The safety definition commonly found in the literature is binary, i.e., the environment is safe if agents do not collide with obstacle regions or each other throughout navigation and vice versa, providing no explicit way to quantify its safety level. For example, two environments may both be collision-free \emph{yet pose different collision risks}, which binary safety definitions cannot distinguish. In this section, we propose a safety metric that quantifies a \emph{scalar} and \emph{continuous} safety level of the environment w.r.t. multi-agent navigation. This metric will serve as our evaluation criterion for the upper-level environment optimization. Specifically, the safety of multi-agent navigation depends both on the spatial constraints of obstacle regions $\{\Delta_j\}_{j=1}^M$ and on the interactions among agents $\{A_i\}_{i=1}^N$. This motivates to define the safety metric from two aspects: \emph{(i)} safety w.r.t. obstacle regions and \emph{(ii)} safety among agents. 

\begin{figure}[t]
	\centering
	\includegraphics[width=0.45\columnwidth]{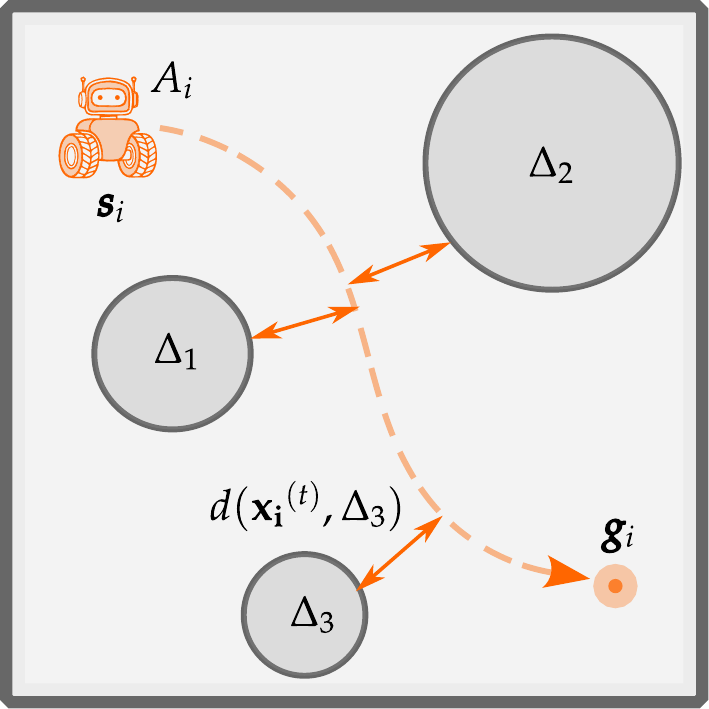}
	\caption{Agent $A_i$ starts from the initial position $\bbs_i$ and moves towards the goal position $\bbg_i$ by following its trajectory $\bbx_i$, and avoids collisions with obstacle regions $\{\Delta_j\}_{j=1}^3$ in the environment $\ccalE$.}
	\label{fig:Obstacle}\vspace{-4mm}
\end{figure}

\subsection{Safety w.r.t. Obstacle Regions}\label{subsec:safeObstacle}

Safety w.r.t. obstacle regions characterizes collision risks of agents from obstacle regions of the environment, which is a direct environmental impact on agents' safety. Specifically, assume agent $A_i$ with trajectory $\bbx_i = [\bbx_i^{(0)}, \ldots, \bbx_i^{(T)}]$ in the environment $\ccalE$ with obstacle regions $\{\Delta_j\}_{j=1}^M$ -- see Fig. \ref{fig:Obstacle} for an illustration. At time step $t$, we quantify its safety w.r.t. obstacle region $\Delta_j$ using a \emph{repulsive potential} function as 
\begin{align}\label{eq:obstacleRPF}
	p_{\Delta_j}(\bbx_i^{(t)}) = 
	\begin{cases} 
		\frac{w}{d(\bbx_i^{(t)}, \Delta_{j}) + \eps}, & \text{if } d(\bbx_i^{(t)}, \Delta_{j}) \le \tau, \\ 
		0, & \text{if } d(\bbx_i^{(t)}, \Delta_{j}) > \tau
	\end{cases}
\end{align}
for $j=1,\ldots,M$, where $d(\bbx_i^{(t)}, \Delta_{j})$ is the shortest distance between the boundaries of $A_i$ and $\Delta_j$, $w$ and $\eps$ are constants of design choice, and $\tau > 0$ is a safety threshold. Here, $\eps > 0$ prevents infinite values when $d(\bbx_i^{(t)}, \Delta_{j}) \to 0$ and $\tau$ represents a safe distance with no collision risk. The value of $p_{\Delta_j}(\bbx_i^{(t)})$ quantifies how close $A_i$ is to $\Delta_j$ and, in turn, characterizes its collision risk w.r.t. $\Delta_j$, i.e., a larger $p_{\Delta_j}(\bbx_i^{(t)})$ indicates a higher risk.\footnote{For invalid trajectories in which $\bbx_i^{(t)}$ lies inside $\Delta_j$, we define $d(\bbx_i^{(t)}, \Delta_{j})$ as the negative of the shortest distance between the boundaries of $A_i$ and $\Delta_j$ to ensure consistency with the safety metric definition, and assume an appropriate $\eps$ such that $d(\bbx_i^{(t)}, \Delta_{j}) \!+\! \eps \!>\! 0$ for all $t$.} Following this rational, we quantify the safety of agent trajectory $\bbx_i$ w.r.t. obstacle regions $\{\Delta_j\}_{j=1}^M$ as
\begin{align}
	\tilde{p}_{\Delta}(\bbx_i) = \frac{1}{M(T+1)} \sum_{t=0}^T 
	\sum_{j=1}^M p_{\Delta_j}(\bbx_i^{(t)}). 
\end{align}
Normalizing $\tilde{p}_{\Delta}(\bbx_i)$ yields $p_{\Delta}(\bbx_i) = p_{\Delta}(\bbx_i) / (w/\eps) \in [0, 1]$, where a larger value represents more collision risks and a lower safety level. We refer to $p_{\Delta}(\bbx_i)$ as the \emph{unsafety mass w.r.t. obstacle regions} of agent $A_i$ in the environment $\ccalE$. 

The \emph{unsafety mass w.r.t. obstacle regions} of the multi-agent system $\ccalA$ is the expected unsafety over all agents, i.e., 
\begin{align}\label{eq:multiSafetyObstacle}
	p_\Delta(\ccalA) \!=\! \frac{1}{N} \sum_{i=1}^N p_\Delta(\bbx_i),
\end{align}
which is also normalized in $[0,1]$. Then, we define the corresponding \emph{safety mass w.r.t. obstacle regions} as 
\begin{align}\label{eq:safetyObstaclesCom}
	s_\Delta(\ccalA) = 1 - p_\Delta(\ccalA). 
\end{align} 
A larger $s_\Delta(\ccalA)$, instead, represents a higher level of safety w.r.t. obstacle regions of $\ccalA$ in $\ccalE$. The pair $\{p_\Delta(\ccalA), s_\Delta(\ccalA)\}$ together provides a complete and unified safety quantification w.r.t. obstacle regions of the environment. 

\begin{figure}[t]
	\centering
	\includegraphics[width=0.45\linewidth]{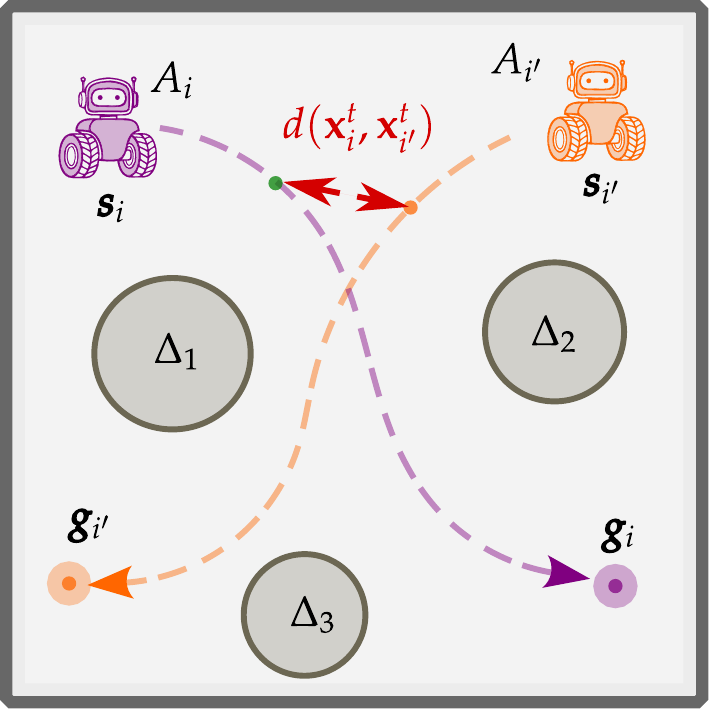}
	\caption{Agents $A_i$, $A_{i^{'}}$ start from initial positions $\bbs_i$, $\bbs_{i^{'}}$ and move towards goal positions $\bbg_i$, $\bbg_{i^{'}}$ by following their trajectories $\bbx_i$, $\bbx_{i^{'}}$, and avoid collisions with obstacle regions $\{\Delta_j\}_{j=1}^3$ and each other.} 
	\label{fig:CollisionZone}\vspace{-4mm}
\end{figure}

\subsection{Safety among Agents}\label{subsec:safeAgent}

Safety among agents characterizes collision risks of agents from other agents, which depends on inter-agent coordination. This is an indirect environmental impact on agents' safety, as their trajectories are generated under spatial constraints of the environment. For example, a well-designed environment may lead to agent trajectories away from each other, while a poorly-designed environment may result in agent trajectories close to or even colliding with each other. Specifically, assume agents $\ccalA = \{A_i\}_{i=1}^N$ with trajectories $\{\bbx_i\}_{i=1}^N$ in the environment $\ccalE$. These agents may pass through the same area at different time steps and come into close proximity, although without direct collisions, i.e., there may exist intersection areas among agent trajectories $\{\bbx_i\}_{i=1}^N$ -- see Fig. \ref{fig:CollisionZone}. We define the collision zone between $A_i$ and $A_{i'}$ as
\begin{align}\label{eq:PotentialCollision}
	\ccalT_{ii'} = \{t ~|~ d(\bbx_i^{(t)}, \bbx_{i'}^{(t)}) \le \tau \}
\end{align}
for $i \ne i' \in \{1,\ldots,N\}$, where $d(\bbx_i^{(t)}, \bbx_{i'}^{(t)})$ is the shortest distance between the boundaries of $A_i$ and $A_{i'}$, and $\tau > 0$ is a safety threshold as in \eqref{eq:obstacleRPF}. The value of $\tau$ can be selected depending on the degree of safety we want to achieve. The collision zone $\ccalT_{ii'}$ assumes that $A_i$ is safe w.r.t. $A_{i'}$ with no collision risk if the distance between two agents at the same time step $t$ exceeds $\tau$. Following the same rational, we quantify the pair-wise safety of $A_i$ w.r.t. $A_{i'}$ at time step $t$ as
\begin{align}\label{eq:SafetyAgent1}
	p_{\bbx_{i'}}(\bbx_i^{(t)}) = 
	\begin{cases} 
		\frac{w}{d(\bbx_i^{(t)}, \bbx_{i'}^{(t)}) + \eps}, & \text{if } t \in \ccalT_{ii'}, \\ 
		0, & \text{if } t \notin \ccalT_{ii'}.
	\end{cases}
\end{align}
This allows to quantify the safety of agent trajectory $\bbx_i$ w.r.t. other agents as 
\begin{align}
	\tilde{p}_{\ccalA}(\bbx_i) = \frac{1}{(N-1)(T+1)} \sum_{i'\ne i} \sum_{t=0}^T p_{\bbx_{i'}}(\bbx_i^{(t)}). 
\end{align}
Normalizing $\tilde{p}_{\ccalA}(\bbx_i)$ yields $p_{\ccalA}(\bbx_i) = \tilde{p}_{\ccalA}(\bbx_i)/({w}/{\eps}) \in [0, 1]$. We refer to $p_{\ccalA}(\bbx_i)$ as the \emph{unsafety mass w.r.t. other agents} of agent $A_i$ in the environment $\ccalE$, where a larger value represents more collision risks and a lower safety level among agents. 

The \emph{unsafety mass among agents} of the multi-agent system $\ccalA$ is averaged over all agents as 
\begin{align}\label{eq:multiSafetyAgent}
	p_\ccalA(\ccalA) = \frac{1}{N} \sum_{i=1}^N p_{\ccalA}(\bbx_i) 
\end{align}
and the corresponding \emph{safety mass among agents} is 
\begin{align}\label{eq:safetyAgentCom}
	s_\ccalA(\ccalA) = 1 - p_\ccalA(\ccalA), 
\end{align} 
which are normalized in $[0, 1]$ as well. A larger $s_\ccalA(\ccalA)$ implies a higher level of safety among agents of $\ccalA$ in $\ccalE$. The pair $\{p_\ccalA(\ccalA),s_\ccalA(\ccalA)\}$ together provides a complete and unified safety quantification among agents in the environment. 

\subsection{A Comprehensive Safety Metric}\label{subsec:properties}

We define the safety metric by combining safety w.r.t. obstacle regions in Section \ref{subsec:safeObstacle} and safety among agents in Section \ref{subsec:safeAgent}. Specifically, define a pair of unsafety and safety masses $\{\ccalP_\ccalE(\ccalA),\ccalS_\ccalE(\ccalA)\}$ by aggregating $\{p_\Delta(\ccalA),s_\Delta(\ccalA)\}$ [cf. \eqref{eq:safetyObstaclesCom}] and $\{p_\ccalA(\ccalA),s_\ccalA(\ccalA)\}$ [cf. \eqref{eq:safetyAgentCom}] with weighted fusion operations as
\begin{equation}\label{eq:safetyTuple}
	\begin{cases}
		\ccalP_\ccalE(\ccalA) = p_\Delta(\ccalA) \cdot M + p_\ccalA(\ccalA) \cdot (N-1), \\
		\ccalS_\ccalE(\ccalA) = s_\Delta(\ccalA) \cdot M + s_\ccalA(\ccalA) \cdot (N-1),
	\end{cases}
\end{equation}
where $M$ represents the number of obstacle regions and $N-1$ represents the number of the other agents w.r.t. each agent itself. The unsafety mass $\ccalP_\ccalE(\ccalA)$ (or the safety mass $\ccalS_\ccalE(\ccalA)$) maps a multi-agent system $\ccalA$ to a real value, which quantifies an explicit safety level of the multi-agent system $\ccalA$ in the environment $\ccalE$. 

\begin{figure}[t]
	\centering
	\includegraphics[width=0.75\linewidth]{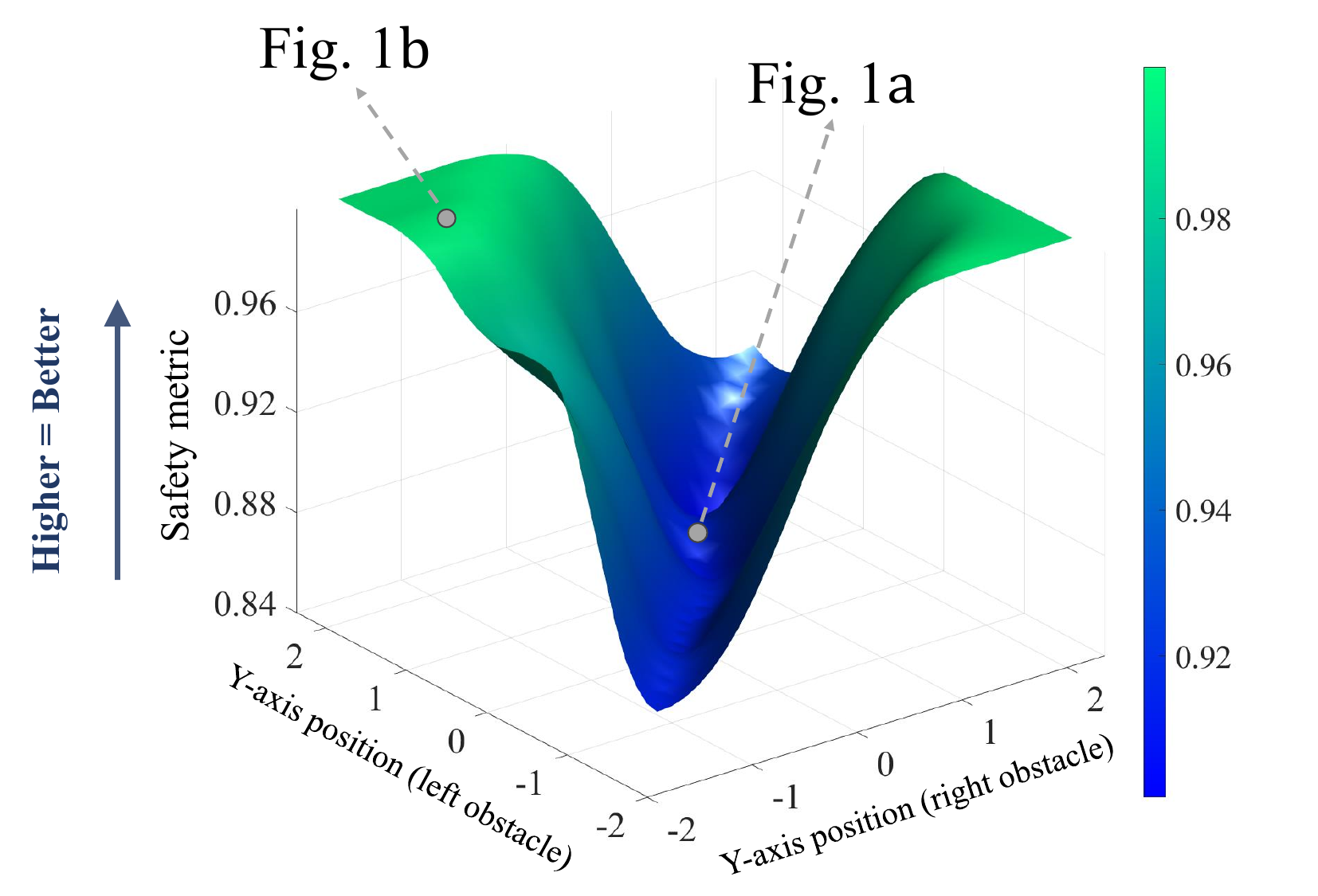}
	\caption{Safety metric as a function of environment parameters in the example environment of Fig. \ref{fig:environment-design-relationship}, where environment parameters are the $y$-axis positions of the left and right obstacle regions.}
	\label{subfig2c}\vspace{-4mm}
\end{figure}

\textbf{Properties.} We show that this is a valid metric satisfying the following properties grounded in measure theory: 

\emph{\textbf{(i)}} 
For any environment $\ccalE$ and multi-agent system $\ccalA$, the unsafety mass is non-negative $\ccalP_\ccalE(\ccalA) \ge 0$.

\begin{figure}%
	\centering
	\begin{subfigure}{0.45\columnwidth}
		\includegraphics[width=\linewidth]{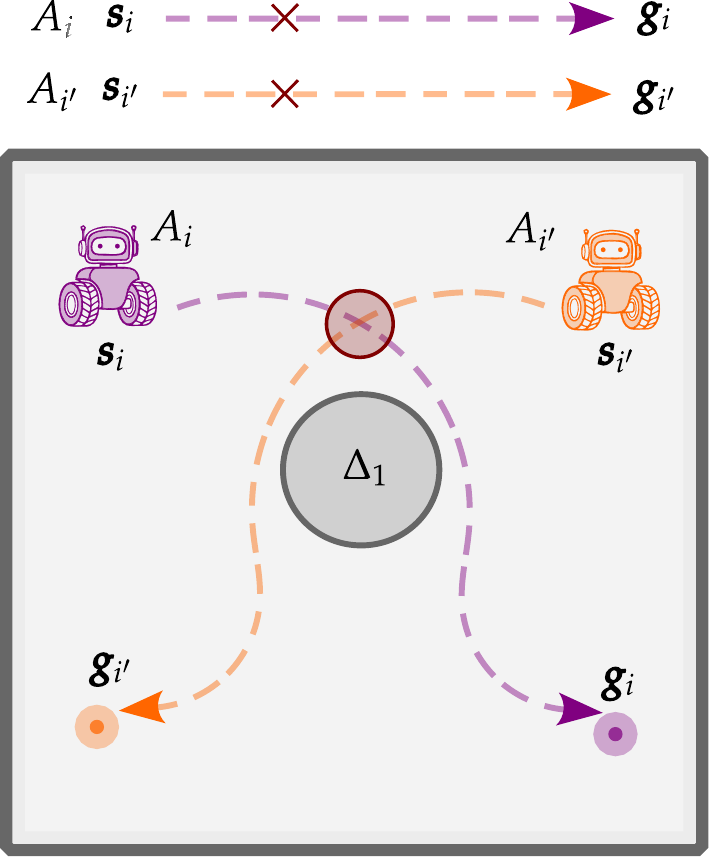}%
		\caption{Poorly-designed environment}%
		\label{subfiga}%
	\end{subfigure}\hfill\hfill%
	\begin{subfigure}{0.45\columnwidth}
		\includegraphics[width=\linewidth]{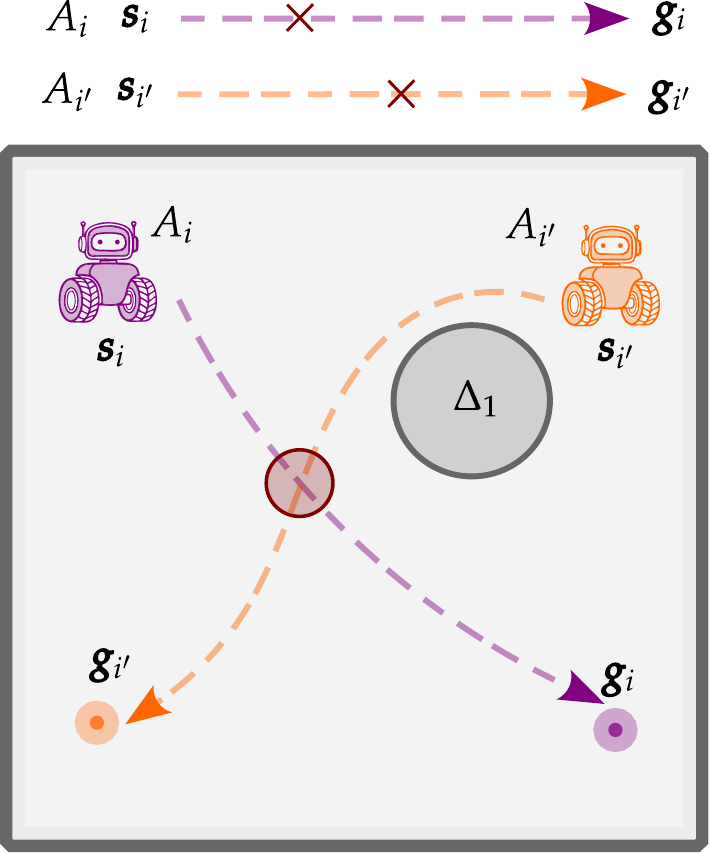}%
		\caption{Well-designed environment}%
		\label{subfigb}%
	\end{subfigure}
	\caption{\textbf{(a)} Multi-agent navigation in a poorly-designed environment. While two agents may navigate from $\bbs_i, \bbs_{i'}$ to $\bbg_i, \bbg_{i'}$ without collisions, there are higher collision risks and they are more likely to collide around the intersection of their trajectories (red circle). \textbf{(b)} Multi-agent navigation in a well-designed environment. By refining the spatial constraint of the obstacle region, it implicitly de-conflicts agents by guiding agent $A_1$ to move faster than $A_2$. This avoids potential collisions of two agents and improves the safety level of the environment, while not compromising their navigation performance.}\label{fig:highExample}\vspace{-4mm}
\end{figure}

\emph{\textbf{(ii)}} 
For any environment $\ccalE$ and two multi-agent systems $\ccalA_1$ with trajectories $\{\bbx_{1, i}\}_{i=1}^N$ and $\ccalA_2$ with trajectories $\{\bbx_{2, i'}\}_{i'=1}^{N'}$, we say $\ccalA_1$ and $\ccalA_2$ are \emph{disjoint} if there is no collision zone between any $\bbx_{1, i}$ of $\ccalA_1$ and $\bbx_{2, i'}$ of $\ccalA_2$ [cf. \eqref{eq:PotentialCollision}]. For a countable collection of pairwise disjoint multi-agent systems $\{\ccalA_k\}_{k=1}^K$ of $\{N_k\}_{k=1}^K$ agents with trajectories $\{\{\bbx_{k, i}\}_{i=1}^{N_k}\}_{k=1}^K$, we have 
\begin{align}
	\Big(\sum_{k=1}^K N_k\Big) \cdot \ccalP_\ccalE(\ccalA) = \sum_{k=1}^K \Big(N_k \cdot \ccalP_\ccalE(\ccalA_k)\Big),
\end{align}
where $\ccalA = \cup_{k=1}^K \ccalA_k$ is the union of $\{\ccalA_k\}_{k=1}^K$.

\emph{\textbf{(iii)}} 
For any environment $\ccalE$ and a countable collection of multi-agent systems $\{\ccalA_k\}_{k=1}^K$, which are not necessarily disjoint, we have
\begin{align}
	  \sum_{k=1}^K \Big(N_k \cdot \ccalP_\ccalE(\ccalA_k)\Big) \le \Big(\sum_{k=1}^K N_k\Big) \cdot \ccalP_\ccalE(\ccalA). 
\end{align}
Proofs of properties \emph{(i)}-\emph{(iii)} are summarized in Appendix \ref{appendix:proofs}.

\begin{figure*}[t]%
	\centering
	\begin{subfigure}[c]{0.15\textwidth}
            \centering
		\includegraphics[width=1\linewidth]{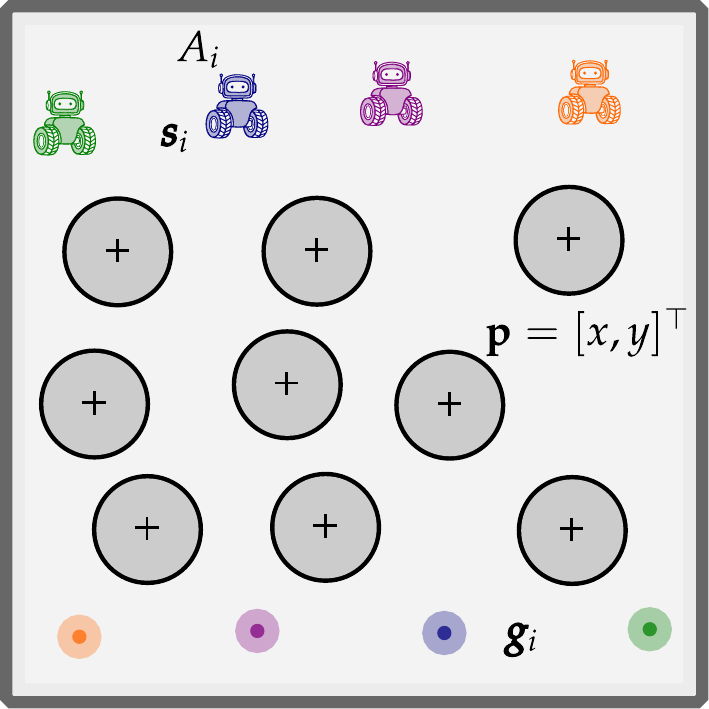}%
		\caption{Warehouse}%
		\label{subfigScenario1}%
	\end{subfigure}
        \hfill
	\begin{subfigure}[c]{0.15\textwidth}
            \centering
		\includegraphics[width=1\linewidth]{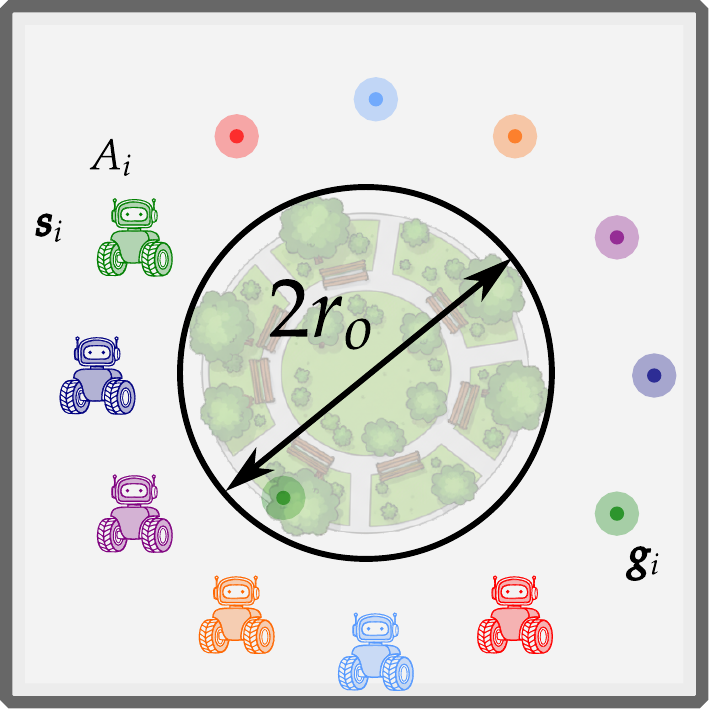}%
		\caption{Roundabout}%
		\label{subfigScenario2}%
	\end{subfigure}
        \hfill
	\begin{subfigure}[c]{0.15\textwidth}
            \centering
		\includegraphics[width=1\linewidth]{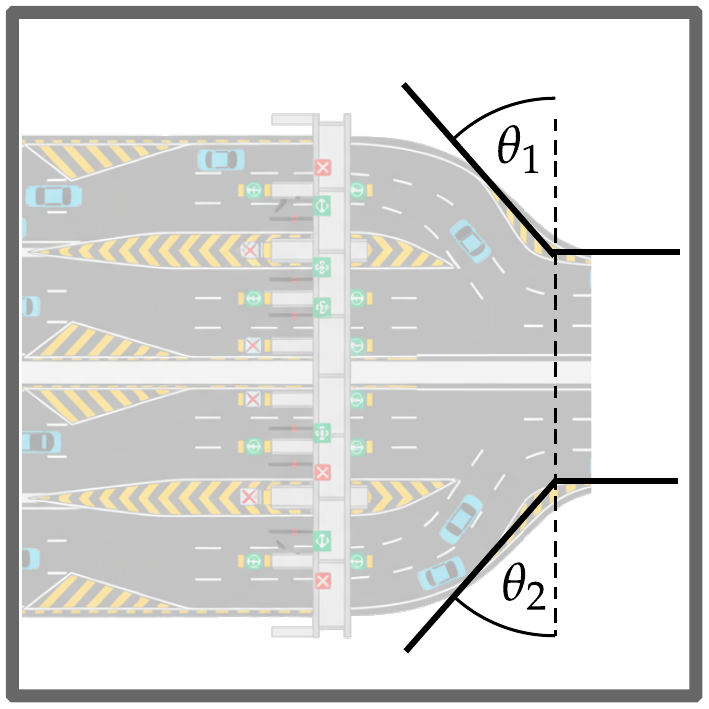}%
		\caption{Narrow passage}%
		\label{subfigScenario3}%
	\end{subfigure}
        \hfill
	\begin{subfigure}[c]{0.15\textwidth}
            \centering
		\includegraphics[width=1\linewidth]{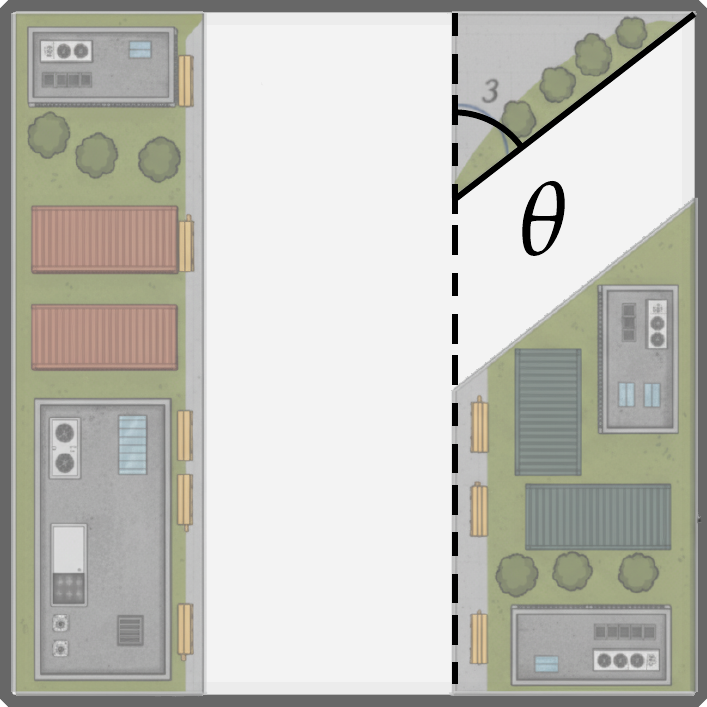}%
		\caption{Highway exit ramp}%
		\label{subfigScenario4}%
	\end{subfigure}
        \hfill
	\begin{subfigure}[c]{0.15\textwidth}
            \centering
		\includegraphics[width=1\linewidth]{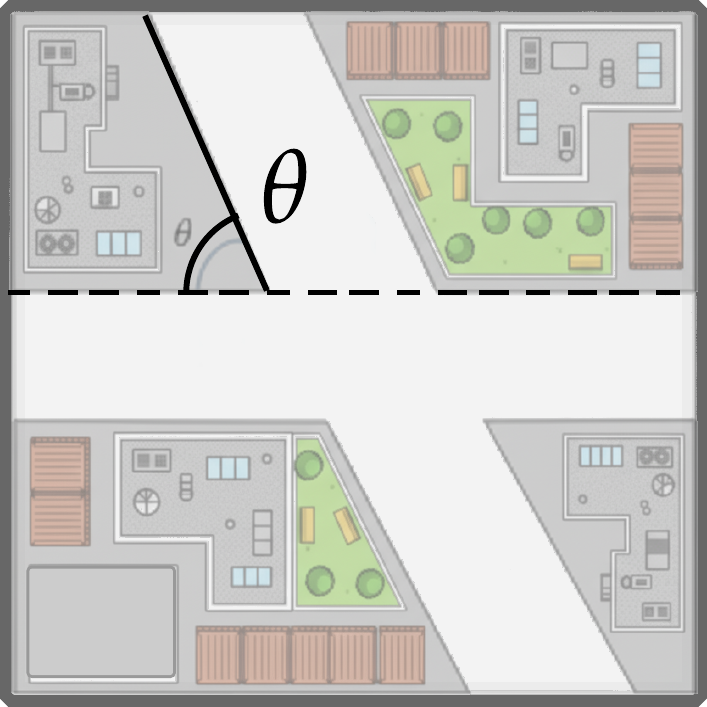}%
		\caption{Road intersection}%
		\label{subfigScenario5}%
	\end{subfigure}\hfill
	\begin{subfigure}[c]{0.15\textwidth}
            \centering
		\includegraphics[width=1\linewidth, trim = {0cm 1.45cm 1.25cm 0.05cm}, clip]{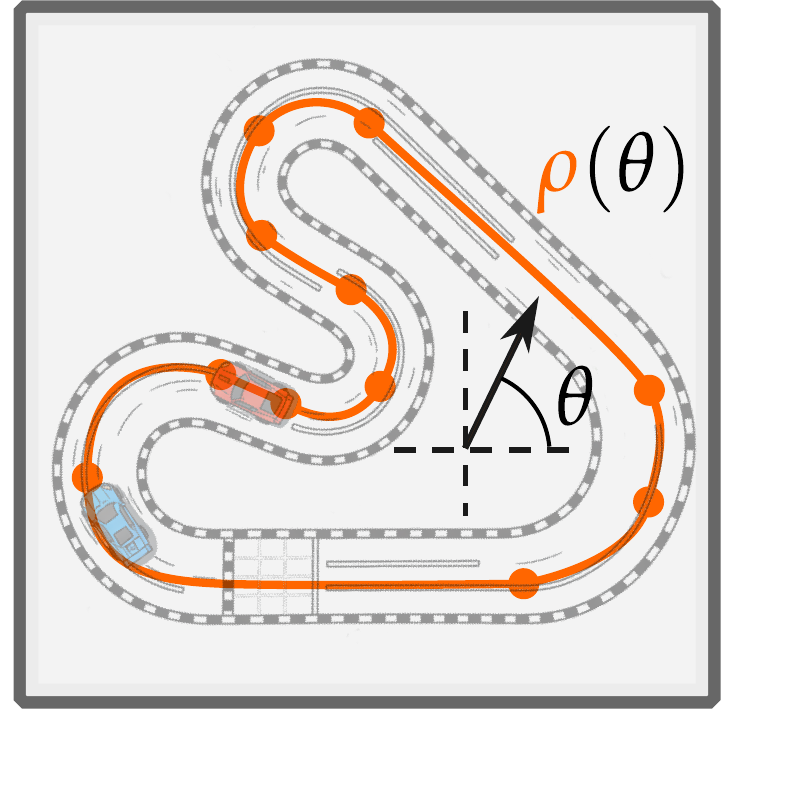}%
		\caption{Track design}%
		\label{subfigScenario6}%
	\end{subfigure}    
	\caption{The six scenarios considered in our experiments. \textbf{(a)} Optimization of shelf positions in a warehouse for safe multi-robot navigation. \textbf{(b)} Optimization of the radius of a roundabout in an urban transportation system for safe multi-vehicle driving. \textbf{(c)} Optimization of narrow passage angles to facilitate vehicles safely passing through the narrow passage. \textbf{(d)} Optimization of the exit angle of a highway to facilitate vehicles safely leaving the highway. \textbf{(e)} Optimization of the intersection angle of two roads for safe multi-vehicle driving. \textbf{(f)} Optimization of the track centerline to facilitate vehicles safely driving on the track.}\label{fig:Scenarios}\vspace{-4mm}
\end{figure*}

Property \emph{(i)} is a basic property for metric definition. Property \emph{(ii)} guarantees that $\ccalP_\ccalE(\ccalA)$ satisfies the fundamental requirement $\ccalP_\ccalE(\ccalA \cup \emptyset) = \ccalP_\ccalE(\ccalA)$, where $\emptyset$ is a null multi-agent system with no agent trajectory. Specifically, for any $\ccalA$ of $N$ agents and a null system $\emptyset$, it holds that
\begin{align}
    &(N\!+\!0) \!\cdot\! \ccalP_\ccalE(\ccalA \cup \emptyset) \!\stackrel{\text{(ii)}}{=}\! N \!\cdot\! \ccalP_\ccalE(\ccalA) \!+\! 0 \!\cdot\!  \ccalP_\ccalE(\emptyset) \!=\! N \!\cdot\! \ccalP_\ccalE(\ccalA) 
\end{align}
and thus, $\ccalP_\ccalE(\ccalA \cup \emptyset) = \ccalP_\ccalE(\ccalA)$. Property \emph{(iii)} follows our intuition that the safety level of the environment is lower, i.e., the environment is less safe, for larger systems with more inter-agent interactions. 

Properties \emph{(i)}-\emph{(iii)} justify that the designed metric is valid to quantify safety levels of environments for multi-agent navigation. As we aim to maximize navigation safety, we set the safety mass $\ccalS_\ccalE(\ccalA)$ as the metric of the upper-level environment optimization in \eqref{eq:EnvOpt}, i.e., 
\begin{align}\label{eq:metricSafetyMass}
	F(\bbx^*(\bbvartheta), \bbu^*(\bbvartheta), \bbvartheta) := R \ccalS_\ccalE(\ccalA),
\end{align}
where $R$ is a regularization constant. This allows us to optimize obstacle regions $\{\Delta_j\}_{j=1}^M$ to maximize the safety level for multi-agent navigation and to ease the task of collision-free trajectory optimization. Fig. \ref{subfig2c} displays how $F(\bbx^*(\bbvartheta), \bbu^*(\bbvartheta), \bbvartheta)$ with $R=1/(M+N-1)$ varies as environment parameters change in the example environment of Fig. \ref{fig:environment-design-relationship}, where the environment parameters are the locations (i.e., Y-axis positions) of two obstacle regions. It follows our intuition that environments with obstacle regions at the same Y-axis position along the diagonal (e.g., Fig. \ref{subfig2a}) create narrow passages, increase collision risks, and have smaller metric values with lower safety levels; environments with obstacle regions at different Y-axis positions (e.g., Fig. \ref{subfig2b}) provide more space to coordinate agents for collision avoidance and have larger metric values with higher safety levels.

\subsection{Discussion}

With the proposed safety metric, the upper-level sub-problem optimizes environment parameters $\bbvartheta$ that refine spatial constraints imposed by obstacle regions to improves the safety level from two aspects: (i) reducing collision risks of agents w.r.t. obstacle regions -- see Section \ref{subsec:safeObstacle}; (ii) providing spatial guidance that implicitly de-conflicts agents to reduce inter-agent collision risks among themselves -- see Section \ref{subsec:safeAgent}. For example, in Fig. \ref{fig:highExample}, a well-designed environment (b) tunes the spatial constraint of the circular obstacle to prioritize and de-conflict agents $A_1$ and $A_2$, which guides $A_1$ moving faster than $A_2$ to avoid their potential collisions, while not compromising navigation performance. Both aspects facilitate agents to generate collision-free trajectories, improving the safety level of the environment. Moreover, the optimized environment $\ccalE^\star$ eases the task of safe multi-agent navigation, because it requires less inter-agent coordination or agent-environment interaction for collision avoidance and allows agents to focus more on their own navigation tasks. 

In this context, the optimized environment not only reduces collision risks, but also implicitly lowers the task difficulty (i.e., computational burden) of trajectory optimization and improves the performance of multi-agent navigation, which we corroborate in our experiments of Section \ref{sec:experiments}.

\begin{remark}
    The proposed method requires computing partial derivatives of the metric function $F$, i.e., the safety metric $\ccalS_\ccalE(\ccalA)$, w.r.t. agent trajectories and environment parameters [cf. \eqref{eq:derivative}]. From the definition of $\ccalS_\ccalE(\ccalA)$, this is equivalent to requiring that the distances $d(\bbx_i^{(t)}, \Delta_j)$ and $d(\bbx_i^{(t)}, \bbx_j^{(t)})$ between agents and obstacles can be expressed by differentiable functions. For example, if obstacle regions are circular obstacles, we can use quadratic functions; if obstacle regions are line boundaries (e.g., traffic roads), we can use perpendicular distance functions. For irregular obstacle regions, we may assume, without loss of generality, that each obstacle can be encircled by an outer circular approximation to compute a conservative distance with a quadratic function. 
\end{remark}

\section{Experiments}\label{sec:experiments}

In this section, we evaluate our differentiable environment-trajectory co-optimization framework in a variety of safe navigation scenarios.

\subsection{Experiment Setup}

We consider six scenarios of safe multi-agent navigation showcased in Fig.~\ref{fig:Scenarios}, which correspond to different real-world applications and are detailed in Section \ref{subsec:performance}. In these scenarios, agents are circular of radius $r_a = 0.3$m and the agent collision avoidance is a quadratic constraint of the form 
\begin{align}
    r_a^2 - \|\bbx_i^{(t)} - \bbx_j^{(t)}\|^2_2 \le 0.
\end{align}
We assume agents with double integrator dynamics in the default setting, and investigate the impact of using different agent dynamics in additional experiments of Appendix \ref{append:addExp}. The results are obtained on a computer with an Intel Core i7-8700 CPU and a NVIDIA GeForce GTX 1080 Ti GPU.

\begin{table*}[h]
\scriptsize
\centering
\caption{Performance of the proposed method and baselines in our scenarios. Larger values of the safety metric or SPL and smaller values of NumCOLL, computation time, PCTSpeed, or distance ratio represent higher safety and efficiency of multi-agent navigation. Mean and standard deviation, where specified, are computed over $20$ random navigation tasks.}
\label{tab:scenarios}
\setlength{\tabcolsep}{3pt} 
\begin{tabular}{llccccc}
\toprule
\textbf{Scenario} &  & \textbf{Safety metric $\uparrow$} & \textbf{SPL $\uparrow$} & \textbf{NumCOLL $\downarrow$} & \textbf{Computation time $\downarrow$} & \textbf{PCTSpeed $\downarrow$} \\
\midrule
\multirow{3}{*}{1-1 Warehouse ($4$ agents)} 
    & Optimized environment (ours) & \textbf{0.932 $\pm$ 0.030} & \textbf{0.948 $\pm$ 0.030} & 0 $\pm$ 0 & \textbf{6.554 $\pm$ 5.135} & \textbf{0.825 $\pm$ 0.035} \\
    & Standard environment with a regular layout & 0.878 $\pm$ 0.049 & 0.930 $\pm$ 0.030 & 0 $\pm$ 0 & 34.384 $\pm$ 41.470 & 0.836 $\pm$ 0.042 \\
    & Baseline environment with a random layout & 0.870 $\pm$ 0.073 & 0.934 $\pm$ 0.022 & 0 $\pm$ 0 & 51.817 $\pm$ 61.533 & 0.838 $\pm$ 0.034 \\
\midrule
\multirow{3}{*}{1-2 Warehouse ($8$ agents)} 
    & Optimized environment (ours) & \textbf{0.944 $\pm$ 0.019} & \textbf{0.944 $\pm$ 0.029} & \textbf{0 $\pm$ 0} & \textbf{40.677 $\pm$ 38.371} & \textbf{0.793 $\pm$ 0.030} \\
    & Standard environment with a regular layout & 0.909 $\pm$ 0.027 & 0.916 $\pm$ 0.019 & 0.063 $\pm$ 0.134 & 93.067 $\pm$ 75.022 & 0.814 $\pm$ 0.022 \\
    & Baseline environment with a random layout & 0.913 $\pm$ 0.030 & 0.909 $\pm$ 0.025 & 0.144 $\pm$ 0.334 & 98.649 $\pm$ 98.703 & 0.822 $\pm$ 0.022 \\
\midrule
\multirow{3}{*}{2 Roundabout} 
    & Optimized environment (ours) & \textbf{0.652} & \textbf{0.891} & 0 & \textbf{25.226} & \textbf{0.727} \\
    & Empty environment with no roundabout & 0.586 & 0.885 & 0 & 45.716 & 0.729 \\
    & Baseline environment with a large roundabout & 0.607 & 0.793 & 0 & 26.123 & 0.733 \\
\midrule
\multirow{3}{*}{3 Narrow Passage} 
    & Optimized environment (ours) & \textbf{0.787} & \textbf{0.979} & 0 & \textbf{184.417} & 0.278 \\
    & Narrow environment & 0.767 & 0.732 & 0 & 236.160 & \textbf{0.248} \\
    & Wide environment & 0.741 & \textbf{0.979} & 0 & 259.853 & 0.276 \\
\midrule
\multirow{3}{*}{4 Highway exit ramp} 
    & Optimized environment (ours) & \textbf{0.860} & \textbf{0.948} & 0 & \textbf{86.437} & \textbf{0.575} \\
    & Baseline with an exit angle $\frac{\pi}{4}$ & 0.797 & 0.928 & 0 & 353.658 & 0.590 \\
    & Baseline with an exit angle $\frac{\pi}{3}$ & 0.808 & 0.941 & 0 & 355.082 & 0.579 \\
\midrule
\multirow{3}{*}{5 Road intersection} 
    & Optimized environment (ours) & \textbf{0.947} & \textbf{0.990} & 0 & \textbf{3.316} & \textbf{0.702} \\
    & Baseline environment with an intersection angle $\frac{\pi}{4}$ & 0.906 & 0.978 & 0 & 15.906 & 0.705 \\
    & Baseline environment with an intersection angle $\frac{3\pi}{4}$ & 0.936 & 0.984 & 0 & 12.361 & 0.707 \\
\midrule
    &  & \textbf{Safety metric $\uparrow$} & \textbf{Distance ratio $\downarrow$} & \textbf{NumCOLL $\downarrow$} & \textbf{Computation time $\downarrow$} & \textbf{PCTSpeed $\downarrow$} \\
\midrule
\multirow{2}{*}{6 Track design} 
    & Optimized environment (ours) & \textbf{0.908} & \textbf{0.952} & 0 & \textbf{7.856} & \textbf{0.574} \\
    & Initial environment & 0.876 & 1 & 0 & 11.300 & 0.604 \\
\bottomrule
\end{tabular}
\end{table*}

We measure the performance w.r.t. four aspects: \emph{(i)} safety metric in Section \ref{sec:Metric} [cf. \eqref{eq:metricSafetyMass}]; \emph{(ii)} average number of collisions (NumCOLL); \emph{(iii)} computation time required by the lower-level trajectory optimization; and \emph{(iv)} Success weighted by Path Length (SPL) and percentage to the maximal speed (PCTSpeed) \cite{anderson2018evaluation}. Specifically, \emph{(i)} measures the safety level of the environment, which indicates how safe an environment is w.r.t. multi-agent navigation and is used to validate the effectiveness of the proposed method. To ease exposition, we normalize the safety metric over obstacles and agents; \emph{(ii)} counts the number of collisions averaged over agents, which results from the fact that trajectory optimization may not find feasible solutions in challenging environments; \emph{(iii)} measures the complexity of the multi-agent navigation task in the environment, which implies how much inter-agent coordination and agent-environment interaction must be handled by trajectory optimization for safe navigation. Both \emph{(ii)} and \emph{(iii)} also correspond to the safety level of the environment, as agents in a safer environment are less likely to collide with obstacle regions or each other and trajectory optimization requires less time to find feasible solutions. For \emph{(iv)}, SPL is the gold standard to measure the efficiency of robot navigation defined as \cite{anderson2018evaluation, gervet2023navigating, wang2019reinforced}
\begin{align}
	\text{SPL} = \frac{1}{N} \sum_{i=1}^N \mathbb{I}_i \frac{\ell_i}{\max\{L_i, \ell_i\}},
\end{align}
where $\mathbb{I}_i$ is a binary indicator for the success of agent $A_i$, $L_i$ is the traveled distance, and $\ell_i$ is the shortest distance. It is a stringent metric that combines success rates with path efficiency. PCTSpeed is the ratio of the average speed to the maximal one, which represents how fast agents move along their trajectories and provides supplementary information for the navigation procedure, complementary to SPL. 

The first three metrics characterize navigation safety in the environment, which is the objective of the upper-level environment optimization and the main focus of this work. A higher value of safety metric [Section \ref{sec:Metric}] or lower values of NumCOLL and computation time represent a higher safety level of the environment and an easier task of collision-free multi-agent navigation. The fourth metric characterizes navigation efficiency in the environment, which is the objective of the lower-level trajectory optimization. From \eqref{eq:ECBF-ZCBF-QP}, a higher SPL represents a higher success rate and improved path efficiency, while a lower PCTSpeed represents lower energy consumption indicating less control effort. These metrics provide a comprehensive evaluation for our method. 

\subsection{Performance Evaluation}\label{subsec:performance}

We evaluate our differentiable bi-level optimization method in the six scenarios of Fig. \ref{fig:Scenarios} and show the results in Table \ref{tab:scenarios}. Additional experiments investigating the impact of agent dynamics and exploring an extension to stochastic co-optimization are presented in Appendix \ref{append:addExp}.

\begin{figure}%
	\centering
	\begin{subfigure}{0.33\columnwidth}
		\includegraphics[width=1\linewidth, height = 0.8\linewidth]{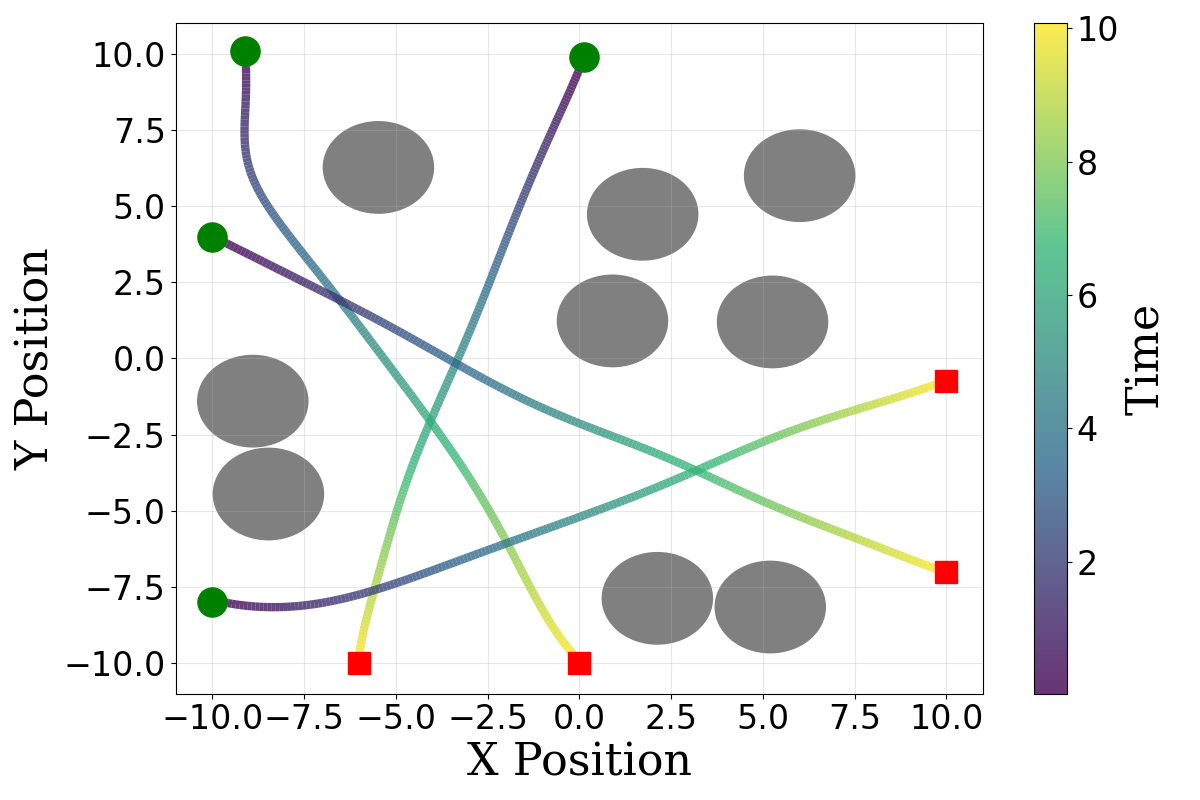}%
		\caption{Optimized}%
		\label{subfigStandardb}%
	\end{subfigure}
	\begin{subfigure}{0.33\columnwidth}
		\includegraphics[width=1\linewidth,height = 0.8\linewidth]{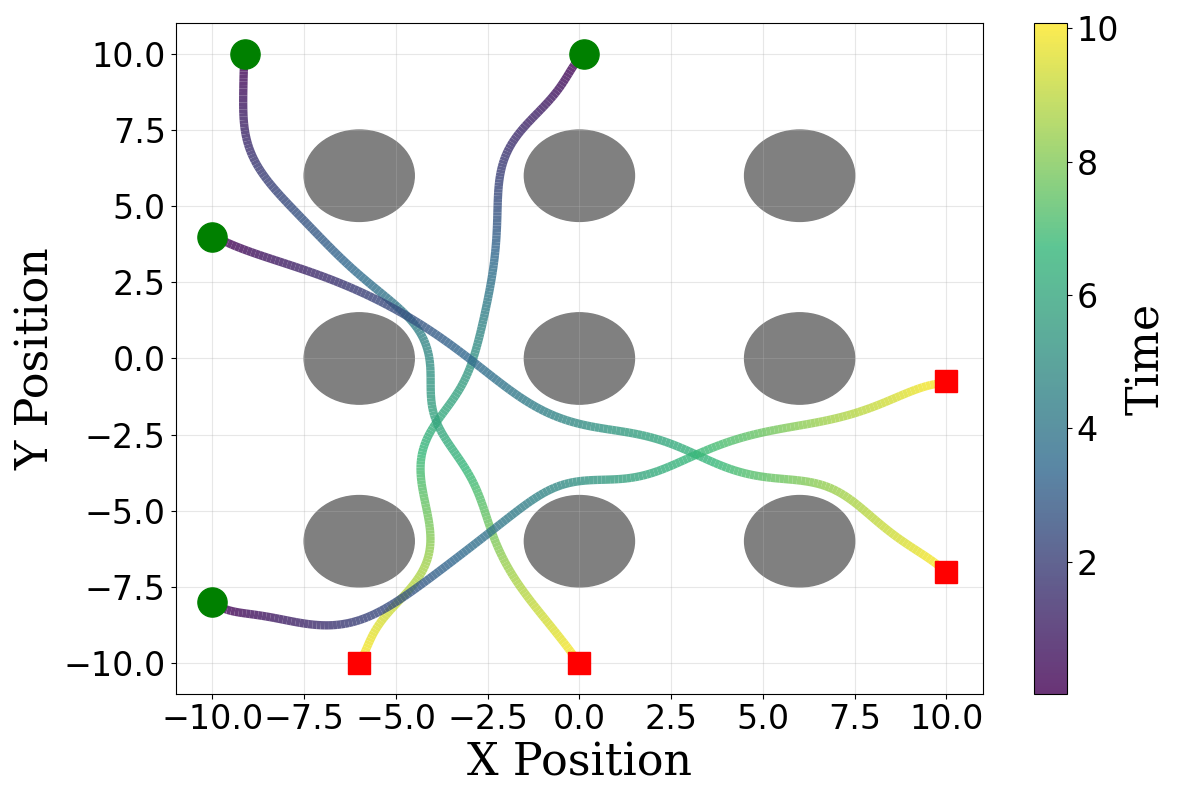}%
		\caption{Standard}%
		\label{subfigStandardc}%
	\end{subfigure}
	\begin{subfigure}{0.33\columnwidth}
		\includegraphics[width=1\linewidth,height = 0.8\linewidth]{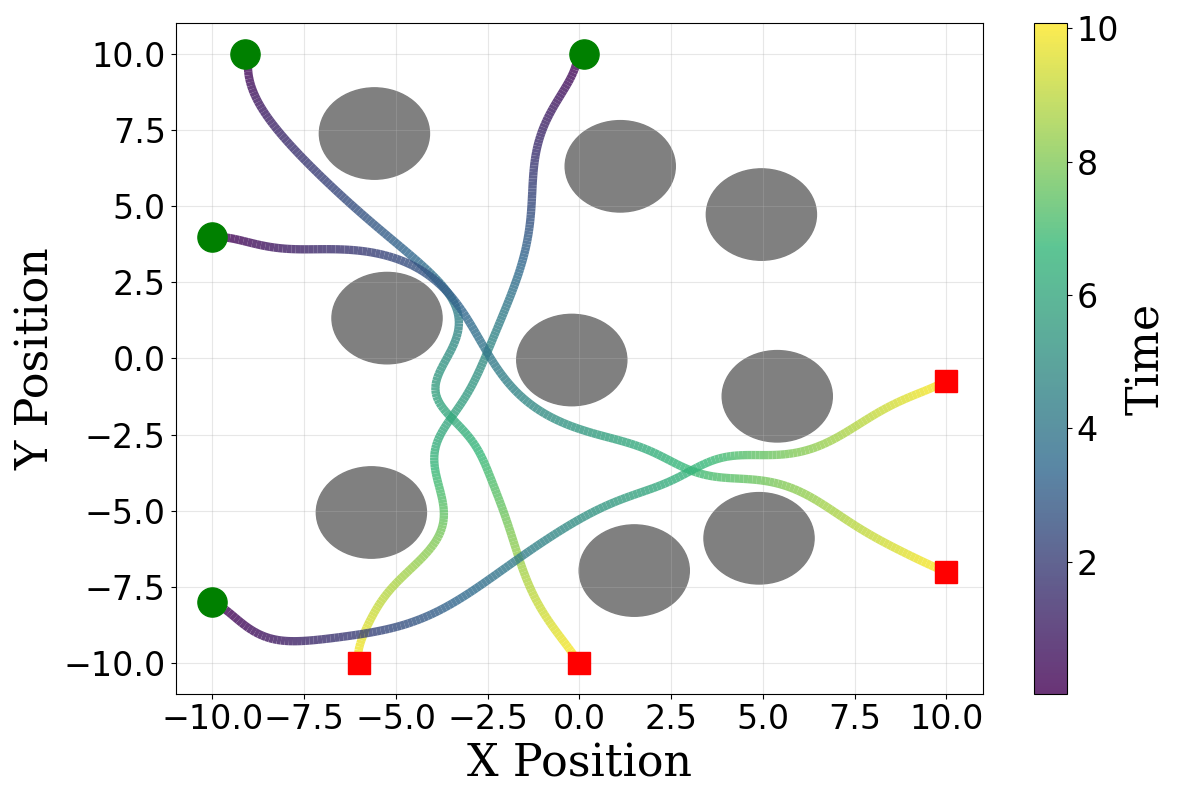}%
		\caption{Random}%
		\label{subfigStandardd}%
	\end{subfigure}
	\caption{ \textbf{(a)} Optimized environment of our method that creates obstacle-free trajectories and prioritizes / de-conflicts agents for safe navigation. \textbf{(b)} Standard environment with a regular shelf layout. \textbf{(c)} Random environment with randomly generated shelf positions.}\label{fig:Standard}\vspace{-4mm}
\end{figure}

\smallskip
\noindent \textbf{Scenario 1 Warehouse.} This scenario considers agents as robots in a warehouse with a set of shelves. The environment is of size $[-10\mathrm{m}, 10\mathrm{m}] \times [-10\mathrm{m}, 10\mathrm{m}]$, and we set $w = 1$, $\gamma = 0.1$ and $\tau = 2.5$m for the safety metric. Four robots are initialized at random positions on one side of the environment and targeted towards goals on the opposite side. The shelves are 9 circular obstacles of radius $r_o=1.5$m, and randomly distributed in the environment -- see Fig. \ref{subfigScenario1} for an illustration. The environment is parametrized by the center positions of the shelves $\{\bbo_j\}_{j=1}^9$, and the constraint of shelf collision avoidance is a quadratic constraint as 
\begin{align}
    \Big(\frac{r_a + r_o}{2}\Big)^2 - \|\bbx_i^{(t)} - \bbo_j\|^2_2 \le 0.
\end{align}
The goal is to optimize the shelf placement to facilitate safe multi-robot navigation. We consider two baselines: (i) a standard environment with a regular shelf layout as in Fig. \ref{subfigStandardc}, which is widely deployed in practice; (ii) a random environment with randomly generated shelf positions. The results are averaged over $20$ random navigation tasks. 

We see that the proposed method outperforms the baselines significantly. \emph{First}, it exhibits the highest safety metric, which indicates that the optimized environment improves the safety level of multi-agent navigation and corroborates the effectiveness of differentiable environment optimization. \emph{Second}, it takes the least computation time to solve the navigation task with trajectory optimization. This is because our method adapts the obstacle layout to provide a safer environment for agents, which reduces collision risks not only between agents and obstacles but also among agents themselves. The latter reduces requirements of inter-agent coordination and agent-obstacle interaction, relaxes constraints of collision avoidance in trajectory optimization, and thus needs less computation time to solve the problem. \emph{Lastly}, it is interesting to find that our method also achieves the highest SPL and the lowest PCTSpeed. This implies that while we aim to optimize the environment to enhance the safety, it implicitly improves the performance as well with higher path efficiency and less control effort, i.e., the objective of trajectory optimization in \eqref{eq:ECBF-ZCBF-QP}. We attribute this behavior to the fact that a safer environment not only reduces collision risks, but also facilitates multi-agent operations and reduces energy consumption. All environments achieve no collision with zero NumCOLL, demonstrating the effectiveness of the lower-level trajectory optimization. 

Figs. \ref{subfigStandardb}-\ref{subfigStandardd} show an example. The optimized environment in Fig. \ref{subfigStandardb}: (i) creates collision-free spaces for agent trajectories to enhance the safety w.r.t. obstacles (Section \ref{subsec:safeObstacle}); (ii) exhibits an irregular structure that prioritizes and de-conflicts the agents to enhance the safety among agents (Section \ref{subsec:safeAgent}). For example, the top-left obstacle is placed in a way that prioritizes the top-right agent in Fig. \ref{subfigStandardb} moving first and the top-left agent moving later, to avoid potential inter-agent collisions. Moreover, the obstacle-free agent trajectories in the optimized environment are path efficient, which are close to the shortest trajectories between starting and goal positions. These aspects together yield enhanced safety and improved performance of multi-agent navigation, compared to the baselines. 

Then, we evaluate our method in a larger system with $8$ agents and $9$ obstacles, where the scenario becomes more cluttered. We similarly observe that our method improves the performance with a higher safety level, less computation time, a larger SPL, and a lower PCTSpeed. The performance improvement introduced by our method increases from smaller to larger multi-agent systems. This indicates that our method is applicable to larger systems, and can provide more benefits in these systems. We also note that the optimized environment maintains no collision with zero NumCOLL, while the baseline environments lead to collisions in some challenging cases. This further validates that our method can ease navigation tasks by improving environment safety, which makes it easier for trajectory optimization to find feasible solutions.

\smallskip
\noindent \textbf{Scenario 2 Roundabout.} This scenario replicates an urban transportation system. We consider agents as vehicles and the obstacle as a roundabout at the intersection of multiple roads. Twelve vehicles are initialized all around the roundabout, which are assumed coming from different roads. They need to cross the roundabout towards the opposite side and drive into the opposite roads, while avoiding collisions with the roundabout and each other. The roundabout is circular and parametrized by the radius $r_o$ -- see Fig. \ref{subfigScenario2}. The goal is to optimize $r_o$ to facilitate safe multi-vehicle driving. We consider two baselines: (i) an empty environment without roundabout, which is often considered ideal with no obstacle hindrance; (ii) an environment with a large roundabout. 

\begin{figure}
	\centering
	\begin{subfigure}{0.33\columnwidth}
		\includegraphics[width=1\linewidth, height = 0.7\linewidth]{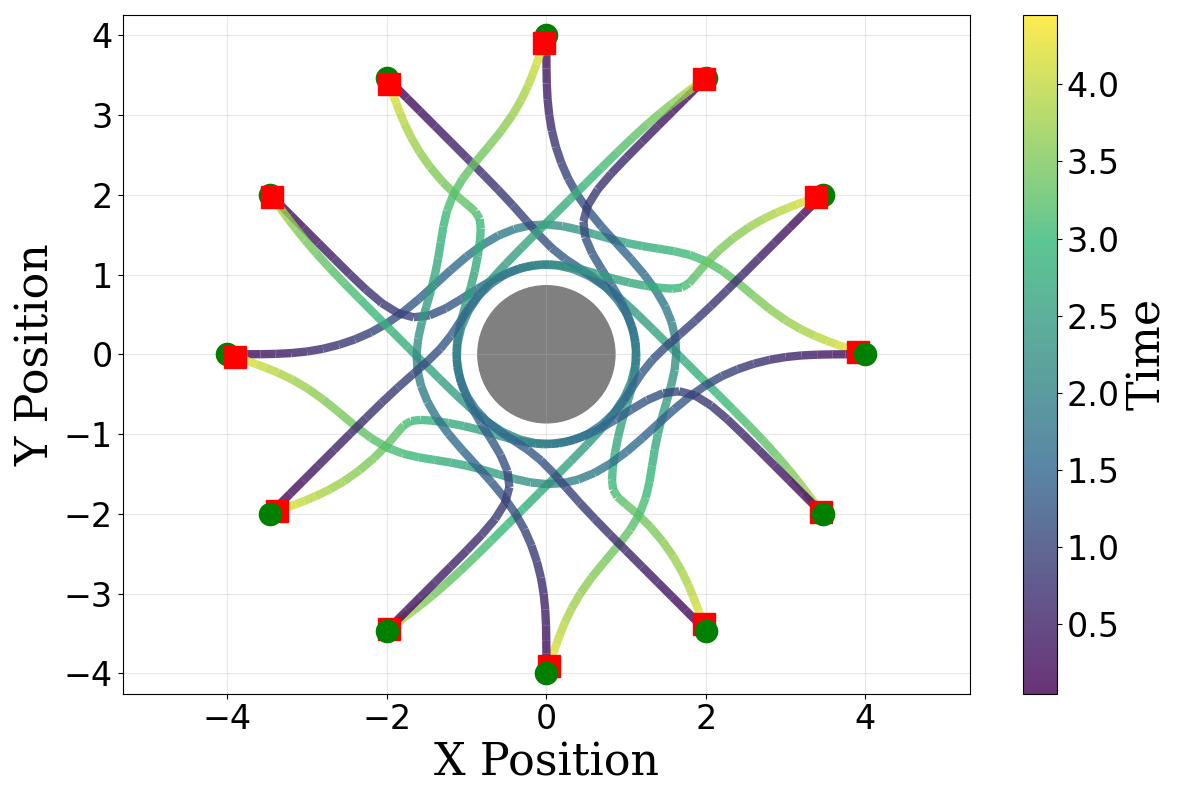}
		\caption{Optimized}
		\label{subfigRoundb}
	\end{subfigure}
	\begin{subfigure}{0.33\columnwidth}
		\includegraphics[width=1\linewidth,height = 0.7\linewidth]{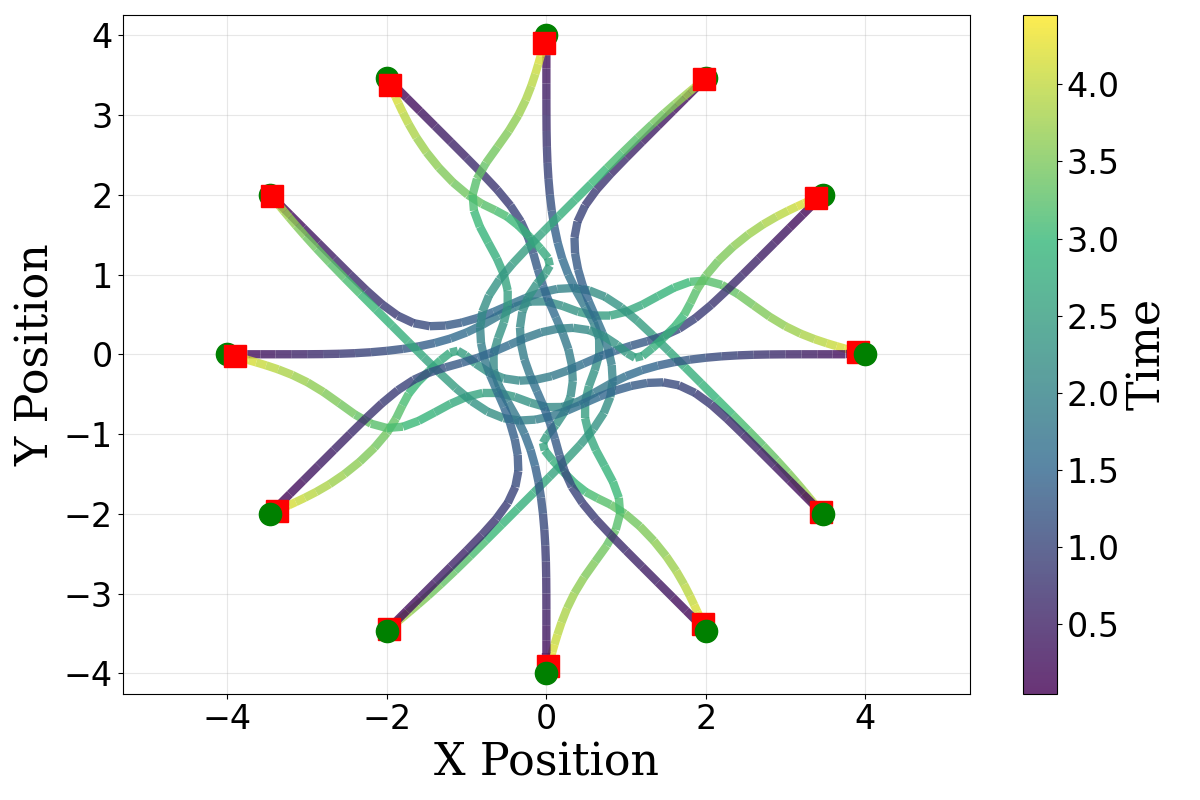}
		\caption{Empty}
		\label{subfigRoundc}
	\end{subfigure}
	\begin{subfigure}{0.33\columnwidth}
		\includegraphics[width=1\linewidth,height = 0.7\linewidth]{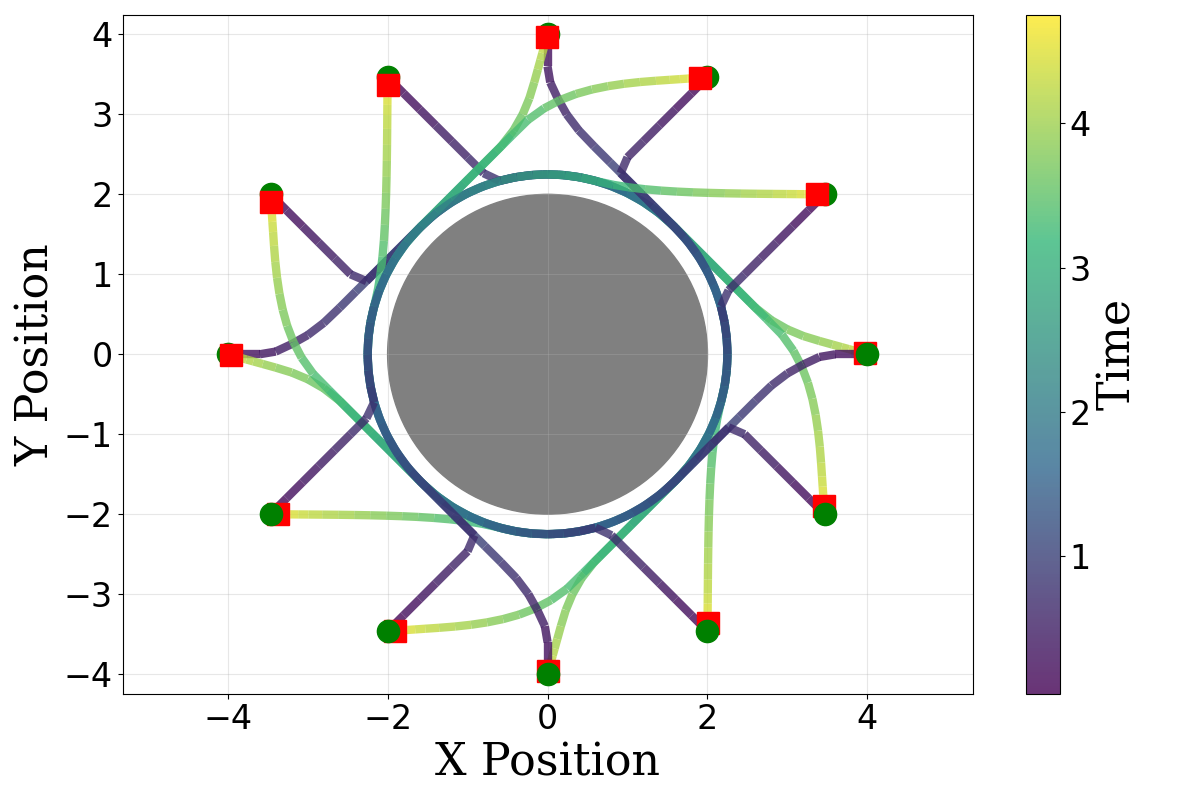}
		\caption{Large}
		\label{subfigRoundd}
	\end{subfigure}
	\caption{\textbf{(a)} Optimized environment that guides agents to move clockwise towards their goal positions. \textbf{(b)} Empty environment with no roundabout that provides no navigation guidance for agents, although imposing no obstacle hindrance. \textbf{(c)} Baseline environment with a large roundabout that blocks efficient pathways, although providing navigation guidance.}\label{fig:roundabout}\vspace{-4mm}
\end{figure}

Table \ref{tab:scenarios} shows the performance, and Figs. \ref{subfigRoundb}-\ref{subfigRoundd} display the optimized environment and the baseline environments. The optimized environment improves the safety level of multi-agent navigation and reduces the computation time of trajectory optimization, while achieving better navigation performance than the baselines. The presence of the roundabout with an appropriate radius provides navigation guidance for agents to move clockwise towards their destinations, which reduces the requirement of inter-agent coordination for collision avoidance and facilitates agent de-confliction for safe navigation. 

The empty environment with no roundabout degrades the safety and increases the computation time significantly, despite no obstacle hindrance along agent trajectories. This is because agents have to coordinate their trajectories by themselves and receive no guidance from the empty environment. The latter makes it challenging for de-confliction, reduces the safety level of multi-agent navigation, and complicates the problem of trajectory optimization, especially in a cluttered case with a large number of agents. The baseline environment with a large roundabout results in a lower SPL with less path efficiency and a higher PCTSpeed with more energy consumption, because it reduces the amount of reachable space and imposes more restrictions on the feasible solution. The latter blocks the shortest pathways and forces agents to move along inefficient trajectories towards destinations. These results corroborate that an appropriate obstacle layout offers guidance and aids in de-conflicting agents for safe navigation, not only hindrance in the traditional view. 

\begin{figure}
	\centering
	\begin{subfigure}{0.33\columnwidth}
		\includegraphics[width=1\linewidth, height = 0.7\linewidth]{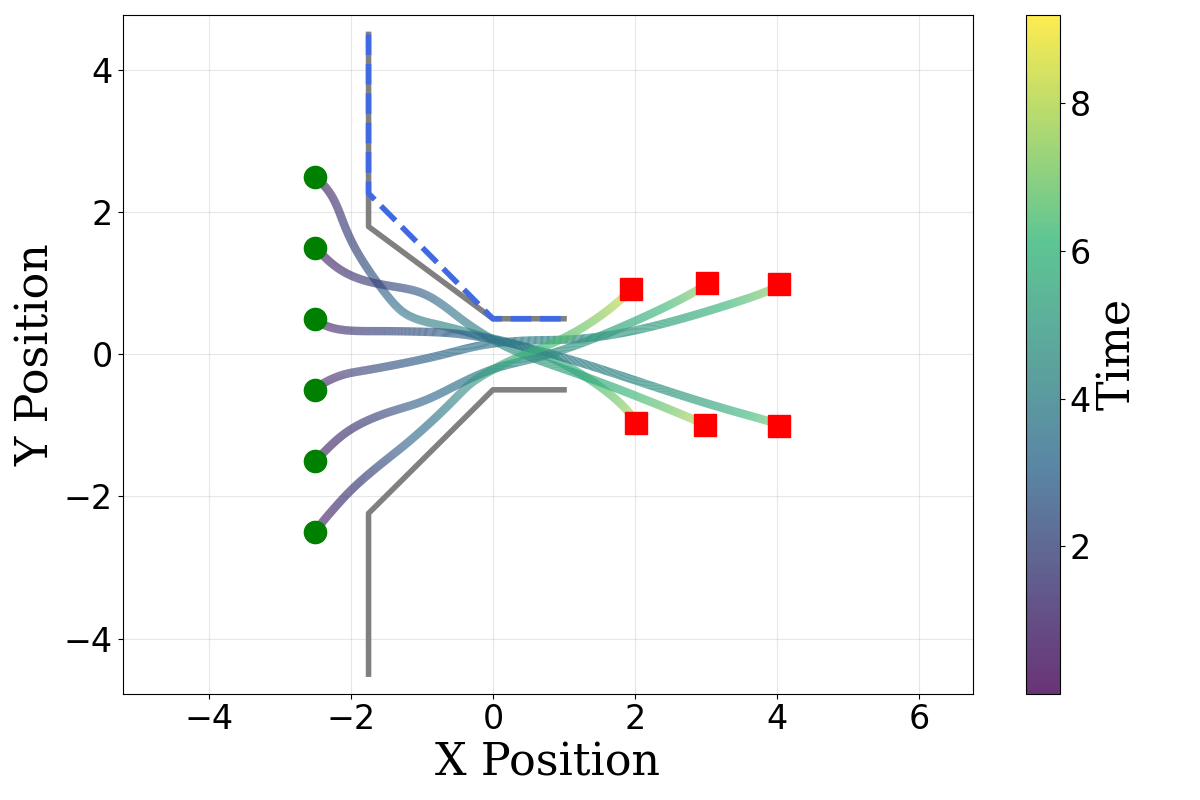}
		\caption{Optimized}
		\label{subfigNarrowb}
	\end{subfigure}\hfill\hfill
	\begin{subfigure}{0.33\columnwidth}
		\includegraphics[width=1\linewidth,height = 0.7\linewidth]{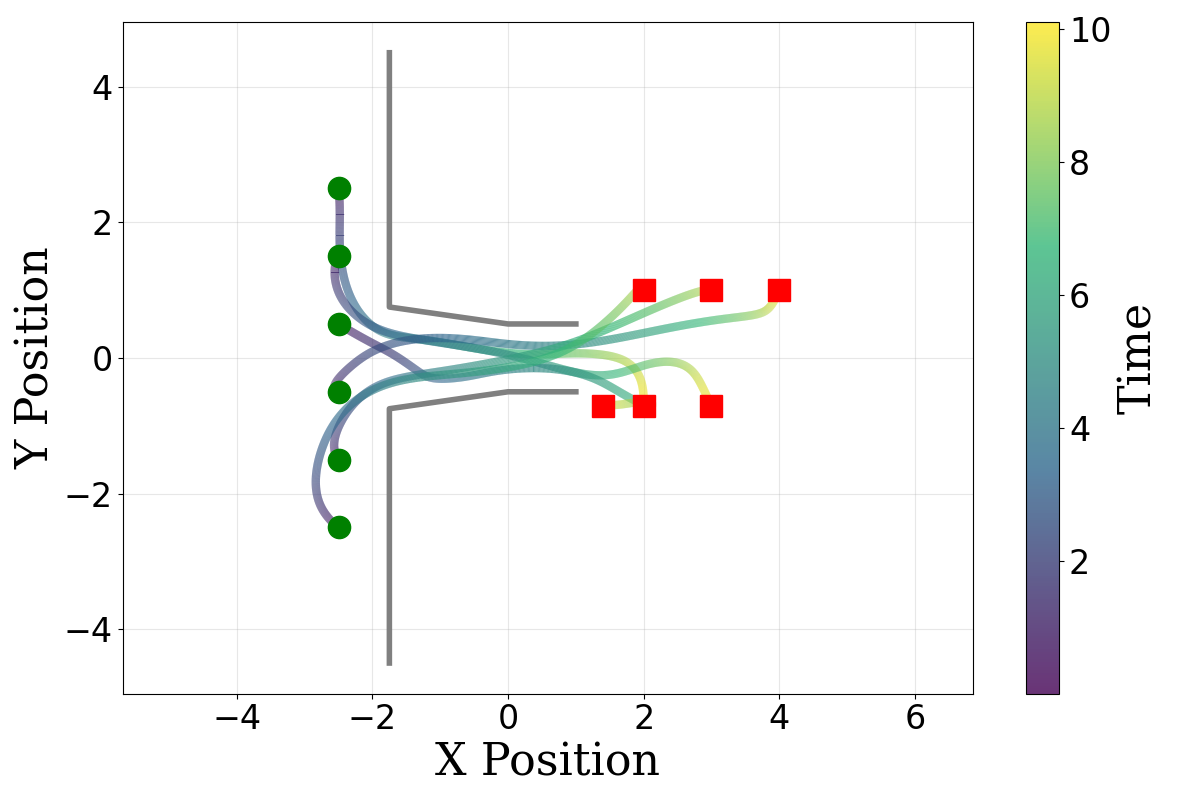}
		\caption{Narrow}
		\label{subfigNarrowc}
	\end{subfigure}\hfill\hfill
	\begin{subfigure}{0.33\columnwidth}
		\includegraphics[width=1\linewidth,height = 0.7\linewidth]{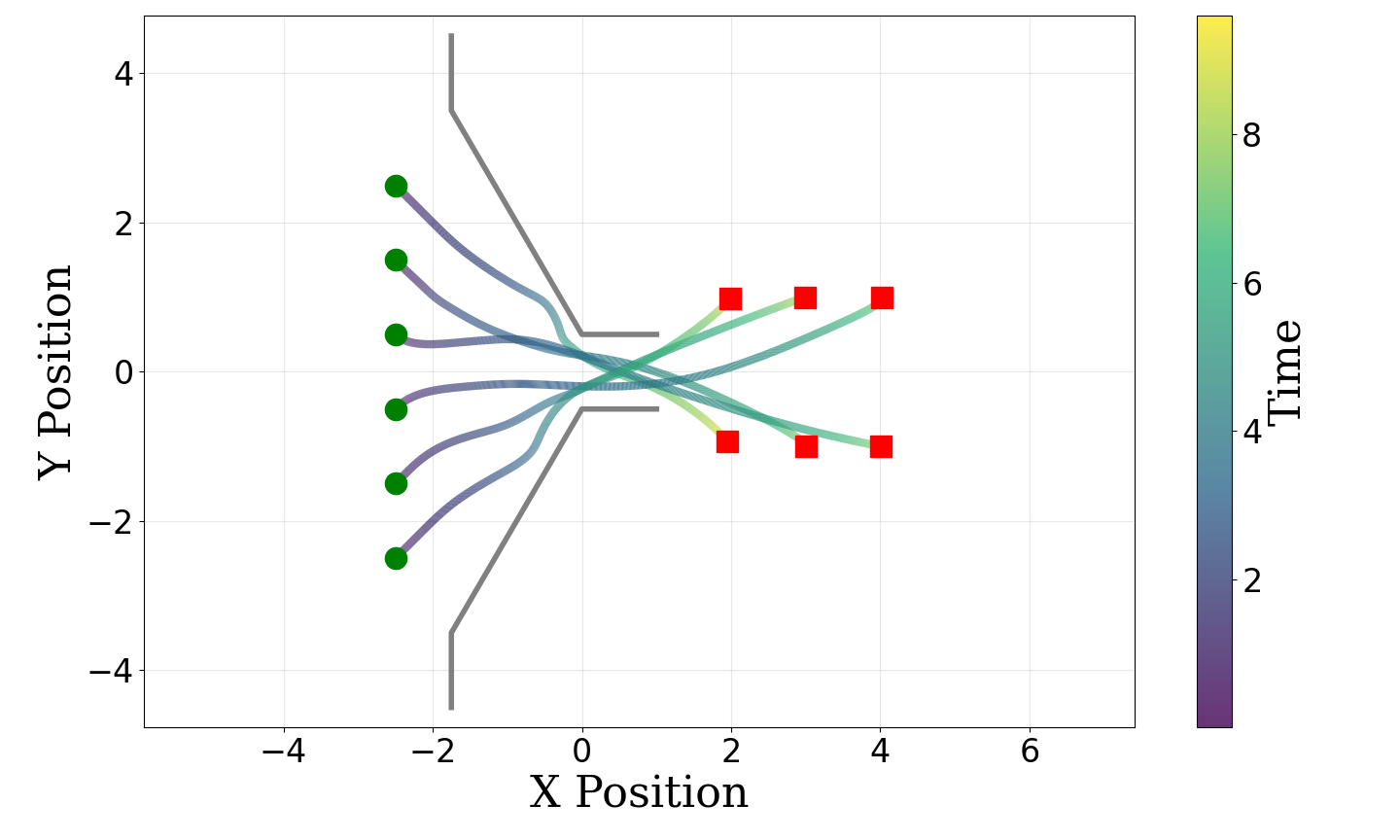}
		\caption{Wide}
		\label{subfigNarrowd}
	\end{subfigure}
	\caption{\textbf{(a)} Optimized environment that creates an asymmetric structure to prioritize / de-conflict agents to pass through the narrow passage. Blue dashed boundaries mirror the lower passage boundaries to demonstrate the asymmetry. \textbf{(b)} Baseline environment with narrow passage angles. \textbf{(c)} Baseline environment with wide passage angles. Both baselines are designed with human intuition and commonly used in practice, while their symmetric structures do not provide de-confliction guidance.}\label{fig:Narrow}\vspace{-4mm}
\end{figure}

\smallskip
\noindent \textbf{Scenario 3 Narrow Passage.} This scenario mimics narrow passages in the real world. Agents are distributed outside a workspace and aim to pass through a narrow passage to get into it. For example, drive vehicles moving towards a toll station in a highway and control ground robots moving into a warehouse through a narrow gate -- see Fig. \ref{subfigScenario3}. As the reachable space and the environment size decrease, we set $\tau = 1$m. The obstacle regions are the workspace boundaries, each expressed as a linear function $R_1 x + R_2 y + R_3 = 0$, and the constraint of obstacle collision avoidance is a quadratic constraint as 
\begin{align}
    \mathbb{I}(\bbx_i^{(t)}) \left[r_a^2 - \frac{(R_1 [\bbx_i^{(t)}]_x + R_2 [\bbx_i^{(t)}]_y + R_3)^2}{R_1^2 + R_2^2}\right] \le 0,
\end{align} 
where $\mathbb{I}(\bbx_i^{(t)})$ is an indicator function that identifies if $\bbx_i^{(t)}$ is in the range of the boundary constraint. The width of the passage is fixed, and the goal is to optimize the passage angles $\theta_1$ and $\theta_2$ that facilitate agents to safely passing through the narrow passage into the workspace. We consider two baselines, where the first has narrow passage angles as in Fig. \ref{subfigNarrowc} and the second has wide passage angles as in Fig. \ref{subfigNarrowd}.

Table \ref{tab:scenarios} shows the results. Our method achieves the highest safety level and the least computation time with comparable navigation performance. Notably, the optimized environment outperforms the baseline environment with wide passage angles, although the wide baseline has the largest amount of reachable space and is generally believed ideal as in common practice. This challenges traditional design assumptions and demonstrates the value of automatic optimization. The baseline environment with narrow passage angles has the least obstacle-free space and imposes the strictest constraints on agent motion. In this context, not all agents complete their navigation tasks, resulting in a significantly lower SPL.  

Figs. \ref{subfigNarrowb}-\ref{subfigNarrowd} show the optimized environment of our method. Different from the baseline environments, it exhibits an asymmetric structure with a narrower upper angle $\theta_1$ and a wider lower angle $\theta_2$. This structural asymmetry prioritizes agents and provides navigation guidance for agent de-confliction. Specifically, the narrower upper angle $\theta_1$ guides the top agent to move more quickly towards the center, allowing it to arrive earlier and take the space to gain priority when traversing the narrow passage; hence, reducing potential collision risks. These aspects improve the safety level and make it easier to solve multi-agent navigation by trajectory optimization.

\smallskip
\noindent \textbf{Scenario 4 Highway.} This scenario is inspired by a deceleration ramp in the real world, which considers agents as vehicles on a highway and obstacle regions as highway boundaries. Vehicles are initialized in parallel lanes on the highway and aim to exit via the ramp, while avoiding collisions with highway boundaries and each other -- see Fig. \ref{subfigScenario4}. The goal is to optimize the exit angle of the ramp $\theta$ that facilitates vehicles to safely leave the highway. We consider two baselines: (i) a highway with an exit angle $\pi / 4$; (ii) a highway with an exit angle $\pi /3$, both following our intuition for good exit designs.

\begin{figure}
	\centering
	\begin{subfigure}{0.33\columnwidth}
		\includegraphics[width=1\linewidth, height = 0.7\linewidth]{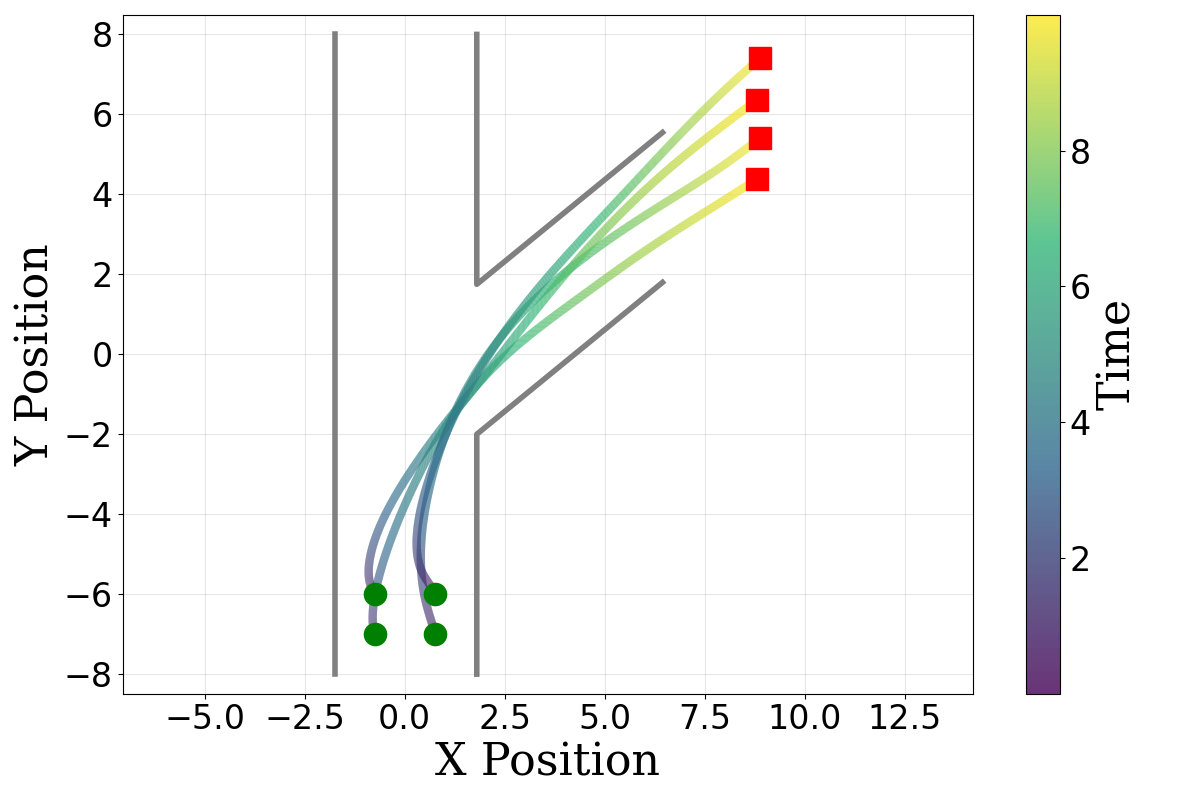}
		\caption{Optimized}
		\label{subfigHighwayb}
	\end{subfigure}
	\begin{subfigure}{0.33\columnwidth}
		\includegraphics[width=1\linewidth,height = 0.7\linewidth]{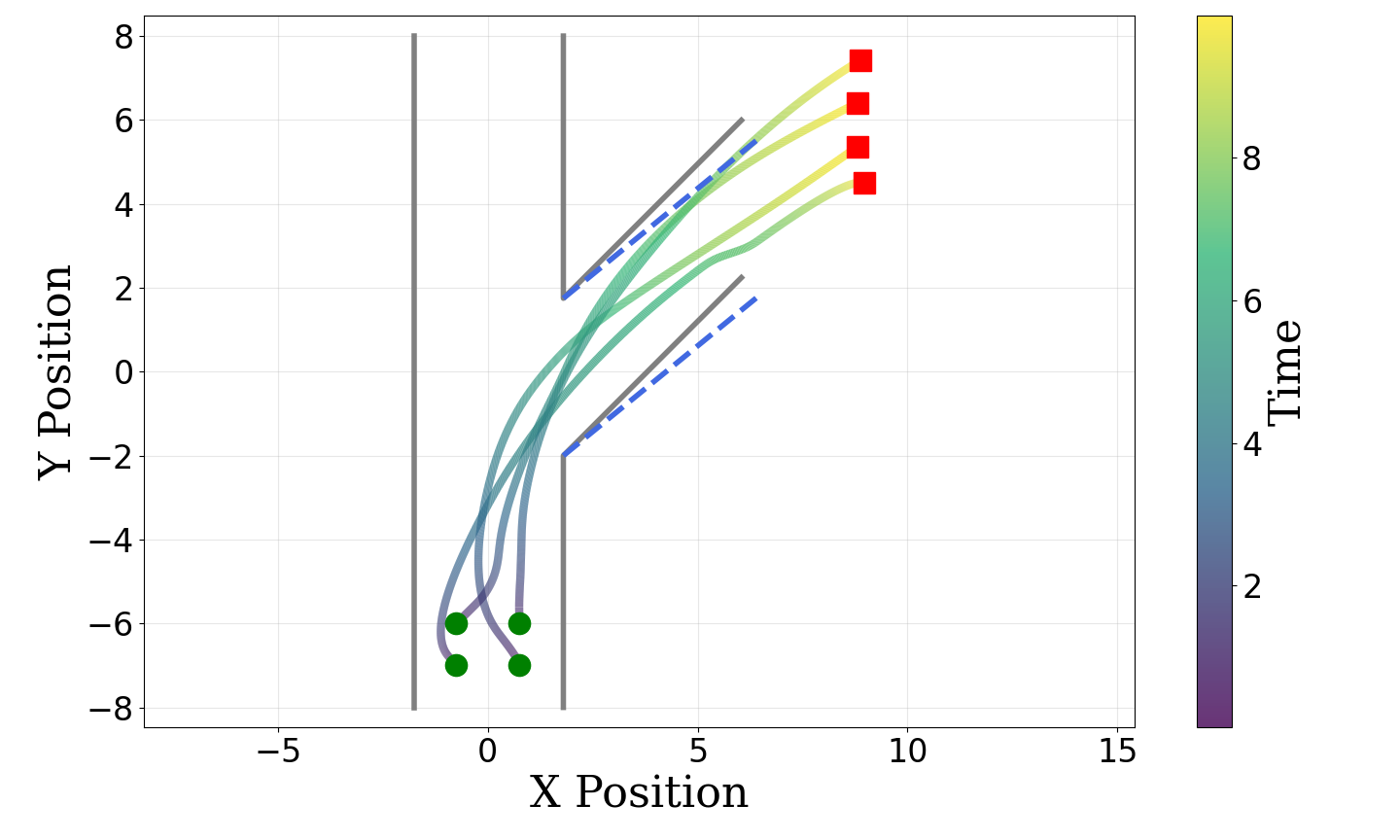}
		\caption{$\pi/4$}
		\label{subfigHighwayc}
	\end{subfigure}
	\begin{subfigure}{0.33\columnwidth}
		\includegraphics[width=1\linewidth,height = 0.7\linewidth]{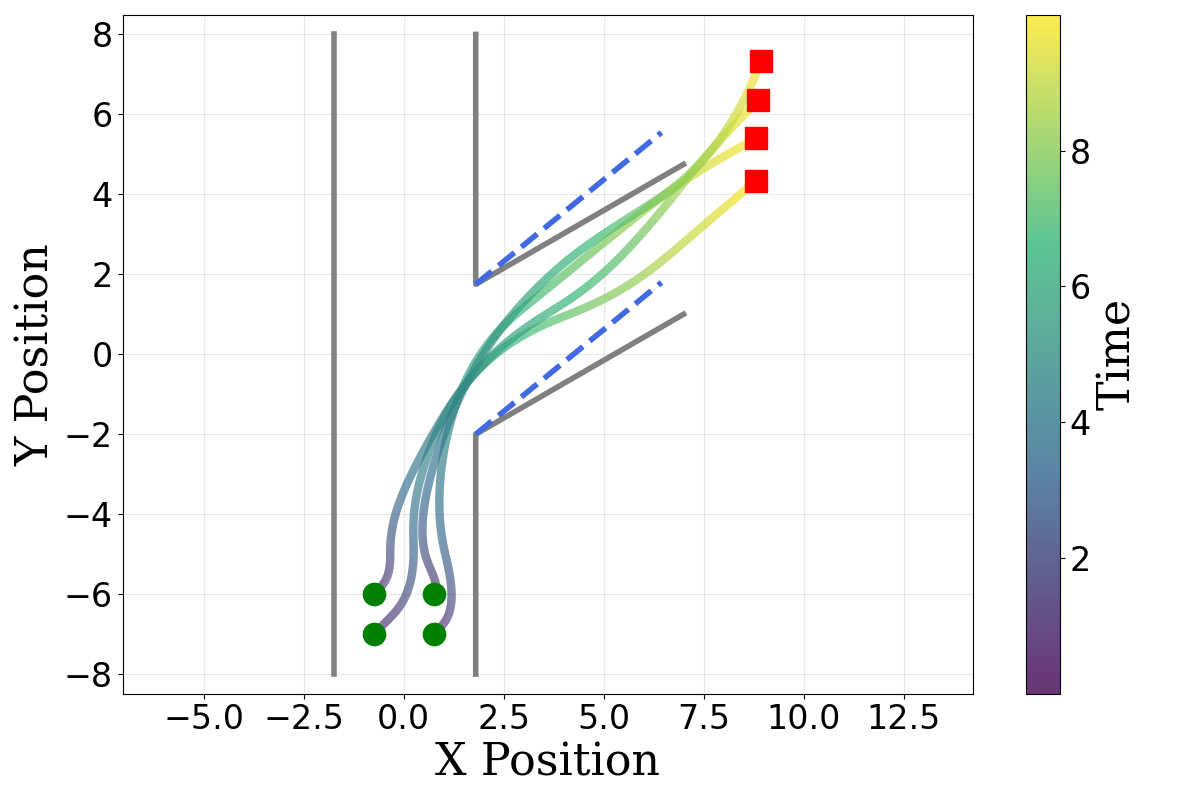}
		\caption{$\pi/3$}
		\label{subfigHighwayd}
	\end{subfigure}
	\caption{\textbf{(a)} Optimized environment that guides agents to leave the highway smoothly towards their goal positions. \textbf{(b)} Baseline environment with an exit angle $\pi/4$. \textbf{(c)} Baseline environment with an exit angle $\pi/3$. Both baselines follow common design practice. Blue dashed boundaries in Figs. \ref{subfigHighwayc}-\ref{subfigHighwayd} represent the optimized environment for reference.}\label{fig:Highway}\vspace{-4mm}
\end{figure}

Table \ref{tab:scenarios} shows the performance. Our method exhibits the highest safety level and requires the least computation time, while delivering the best navigation performance with the highest SPL and the lowest PCTSpeed. The optimized exit angle differs from the intuitive $\pi/4$ and $\pi/3$ baselines, which demonstrates the challenge of manually designing environments in the continuous domain, corroborates the necessity of environment optimization, and validates the effectiveness of the proposed differentiable method. Figs. \ref{subfigHighwayb}-\ref{subfigHighwayd} illustrate the optimized environment and the baseline environments. The optimized highway guides agents to exit smoothly, reducing collision risks either with highway boundaries or among agents. While not significantly changing the exit angle, it effectively improves the safety level and reduces the computation time. This indicates that we can facilitate safe and efficient navigation with no need of large environment changes.

\smallskip
\noindent \textbf{Scenario 5 Road Intersection.} This scenario aims to capture real-world urban traffic, which considers agents as vehicles and obstacle regions as road boundaries. Vehicles are initialized at two different but intersecting roads. They aim to pass through the intersection to keep moving along their roads, while avoiding collisions with road boundaries and each other -- see Fig. \ref{subfigScenario5}. The goal is to optimize the intersection angle $\theta$ that facilitates safe multi-vehicle driving.\footnote{Following real-world settings, as initial and goal positions of vehicles are distributed on roads and roads rotate with the intersection angle $\theta$, initial and goal positions change with $\theta$ as well.}  We consider two baselines: (i) two roads with an intersection angle $\pi / 4$; (ii) two roads with an intersection angle $3\pi / 4$. The first follows our intuition that more parallel roads may reduce agent conflicts and help collision avoidance, while the second is the opposite case for comparison. The results are shown in Table \ref{tab:scenarios}.

\begin{figure}
	\centering
	\begin{subfigure}{0.33\columnwidth}
		\includegraphics[width=1\linewidth, height = 0.7\linewidth]{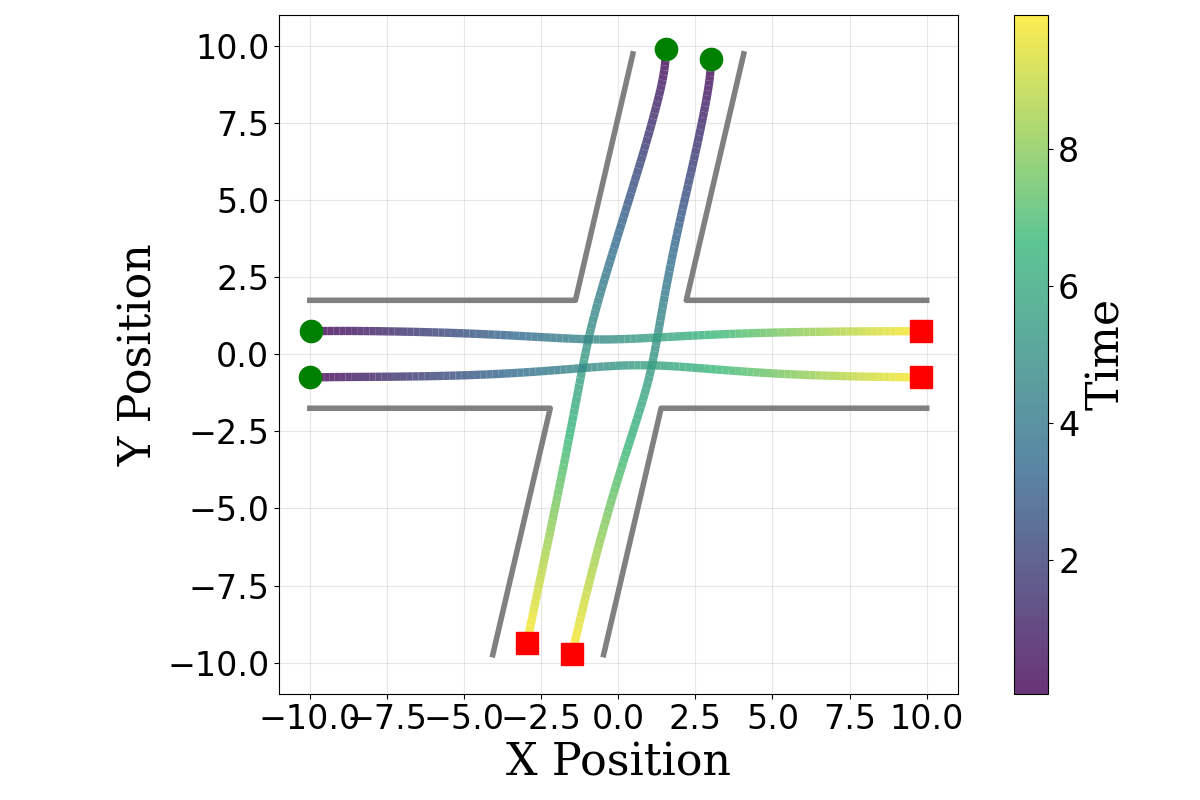}
		\caption{Optimized}
		\label{subfigInterb}
	\end{subfigure}\hfill\hfill
	\begin{subfigure}{0.33\columnwidth}
		\includegraphics[width=1\linewidth,height = 0.7\linewidth]{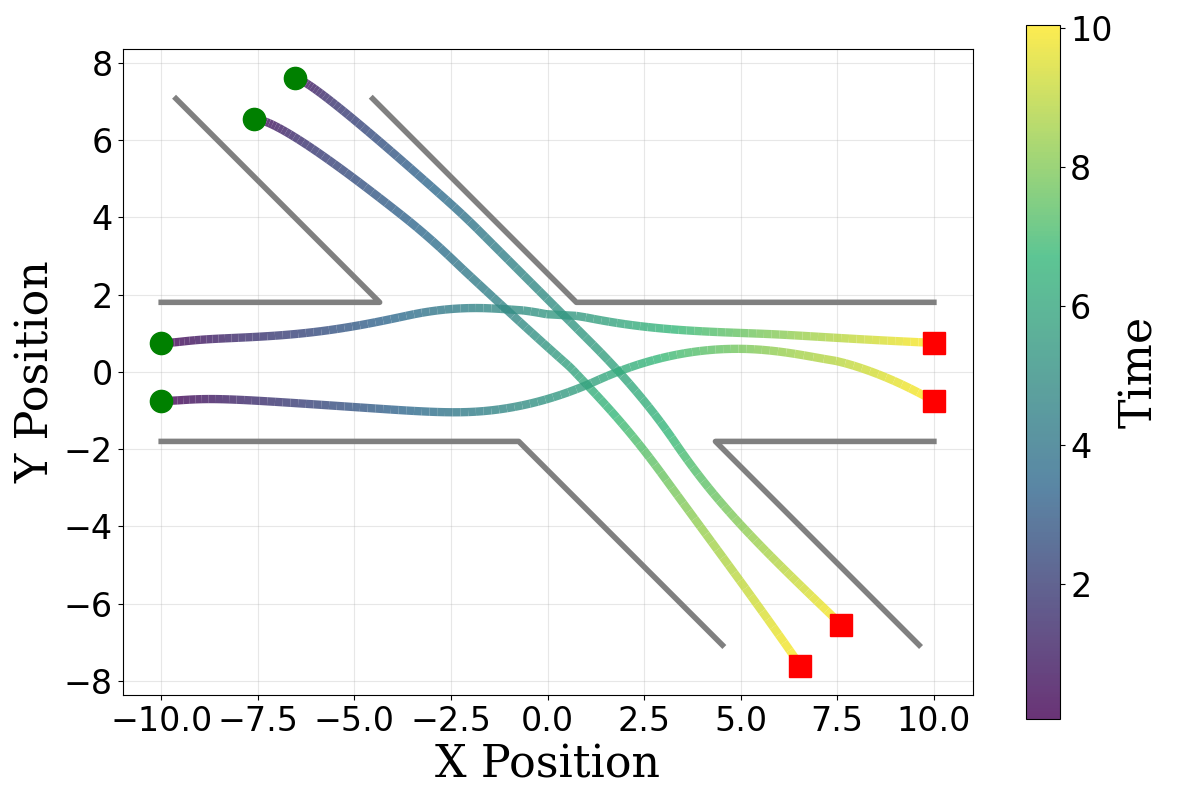}
		\caption{$\pi/4$}
		\label{subfigInterc}
	\end{subfigure}\hfill\hfill
	\begin{subfigure}{0.33\columnwidth}
		\includegraphics[width=1\linewidth,height = 0.7\linewidth]{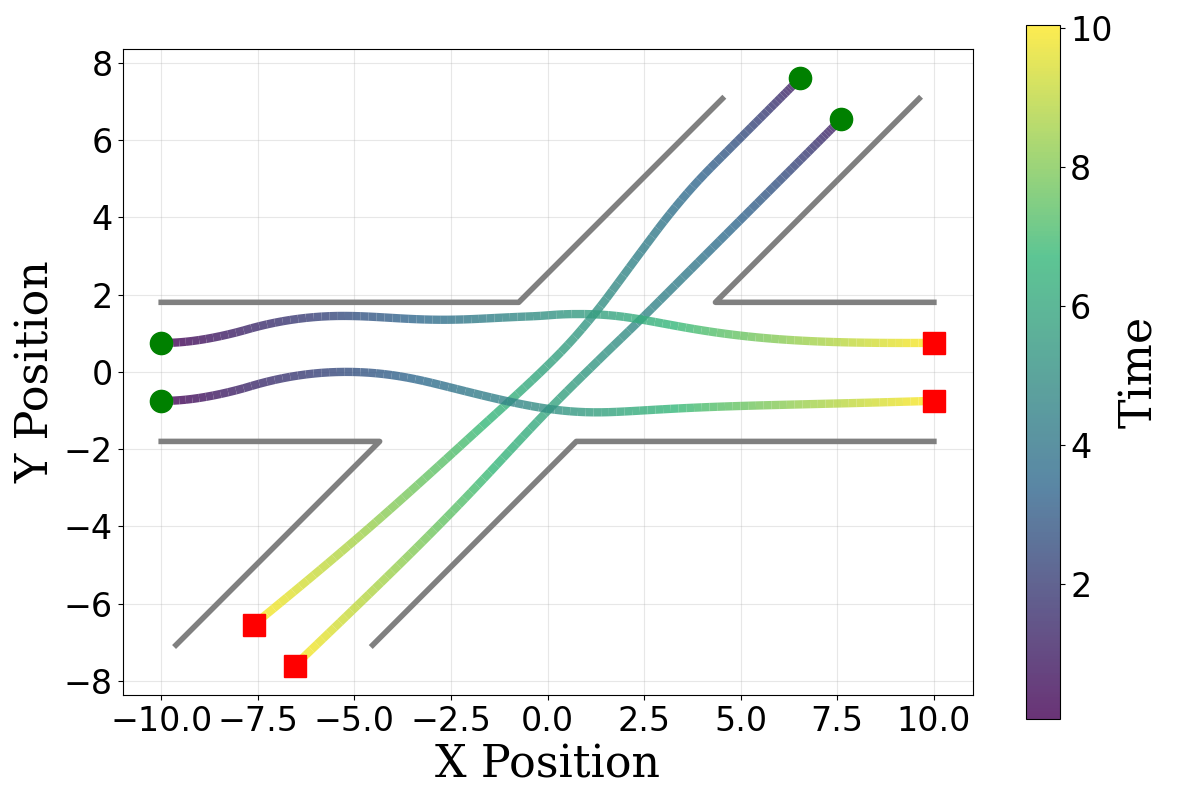}
		\caption{$3\pi/4$}
		\label{subfigInterd}
	\end{subfigure}
	\caption{\textbf{(a)} Optimized environment that guides agents to pass through the intersection smoothly towards their goal positions. \textbf{(b)} Baseline environment with an intersection angle $\pi/4$. \textbf{(c)} Baseline environment with an intersection angle $3\pi/4$.}\label{fig:Inter}\vspace{-4mm}
\end{figure}

The proposed method outperforms the baselines, achieving the highest safety level, the lowest computation time, and the best navigation performance. This similarly indicates that an appropriate intersection angle provides guidance for agents to enhance safety and facilitate navigation, which correspond to Scenarios 1-4 and corroborate the effectiveness of our method. Figs. \ref{subfigInterb}-\ref{subfigInterd} display the optimized environment and the baseline environments. The optimal intersection angle is larger than expected, challenging the intuition that smaller angles, i.e., more parallel roads, reduce agent conflicts and benefit safe navigation. This highlights the difficulty of hand-designing environments and the significance of automatic optimization.

\begin{figure}
    \centering
    \includegraphics[width=0.45\linewidth, trim = {5cm 5cm 0cm 0cm}, clip]{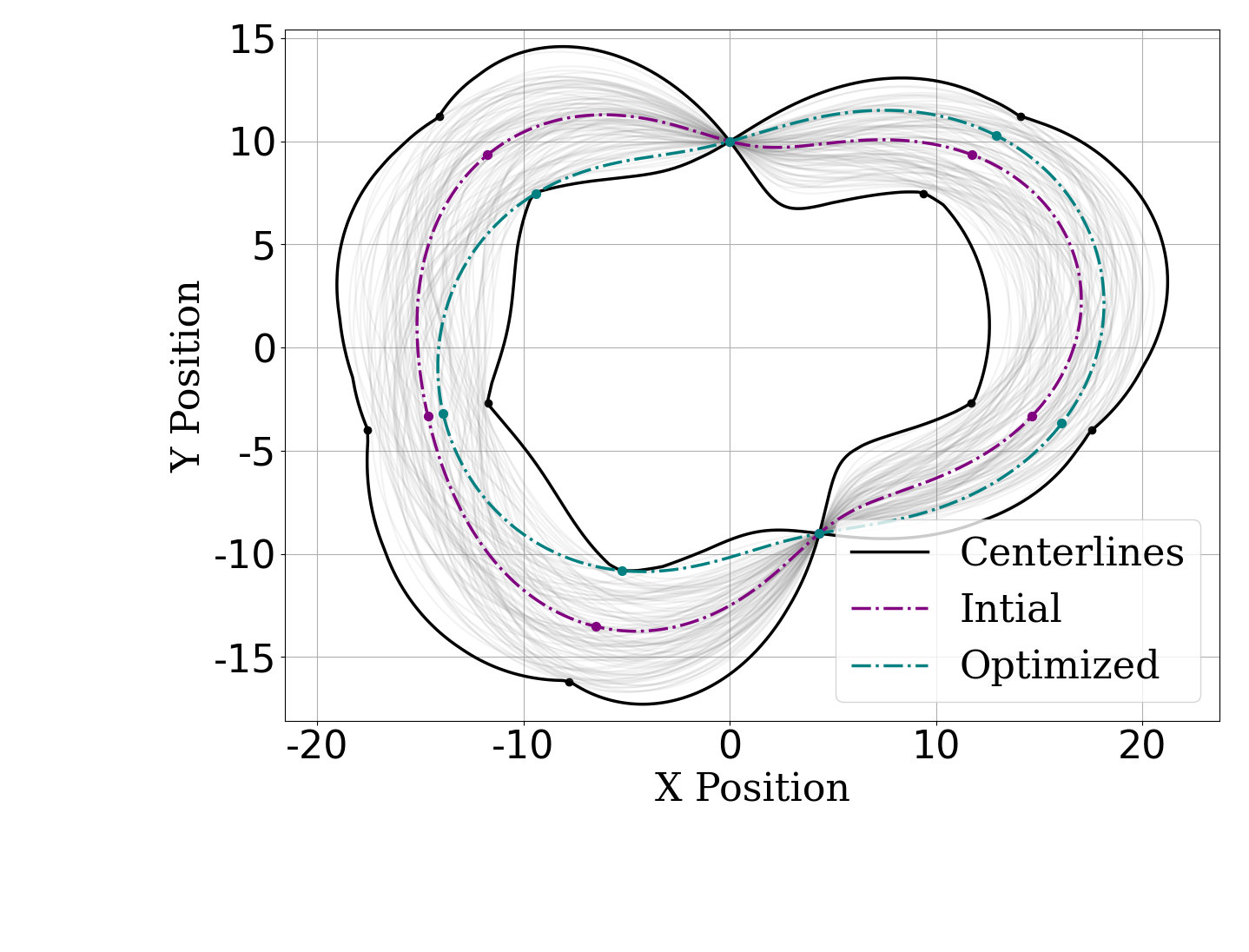}
    \caption{Approximation of the ensemble of possible tracks' centerlines. In grey $50$ random track designs are selected by picking points within the boundaries of the minumum and maximum black control points at equally spaced angular locations. We also showcase the initial and optimized track designs for reference.} 
    \label{fig:track_designs}\vspace{-4mm}
\end{figure}

\smallskip
\noindent \textbf{Scenario 6 Track design.} This scenario involves the design of a track, which considers agents as vehicles and obstacle regions as track boundaries. The track is built by offsetting the centerline with a half-width of $1.25$m to form 
track boundaries. Vehicles are initialized in different lanes of the track and aim to navigate the track in opposite directions. The task is to complete one loop, while avoiding collisions with track boundaries and each other. The agents' states (positions and velocities) and actions (accelerations) are expressed in polar coordinates ($\rho, \theta$) as
\begin{align}
    \mathbf{x}_{i}^{(t)} =[\rho_{i}^{(t)},\theta_{i}^{(t)},\dot\rho_{i}^{(t)},\dot\theta_{i}^{(t)}]^\top,~\mathbf{u}_{i}^{(t)} =[\ddot\rho_{i}^{(t)},\ddot\theta_{i}^{(t)}]^\top. 
\end{align}
The position of agent $A_i$ can be recovered in Cartesian space as 
$\mathbf{p}_i^{(t)} = [\rho_i^{(t)} \cos(\theta_i^{(t)}), \; \rho_i^{(t)} \sin(\theta_i^{(t)})]^{\top}$. 
The environment is parameterized by seven waypoints equally distributed in $[0, 2 \pi]$, which construct a centerline following trigonometric interpolation that ensures periodicity. We consider that the radii of two waypoints are fixed, while those of the other waypoints are reconfigurable within $\pm 20\%$ of the initial track. 
A visualization of the envelope of all possible centerlines is shown in Fig.~\ref{fig:track_designs}. The goal is to optimize the radii of reconfigurable waypoints to facilitate safe multi-vehicle driving. 

\begin{figure}
	\centering
	\begin{subfigure}{0.475\linewidth}
		\centering
		\includegraphics[width=1\linewidth, height = 0.7\linewidth]{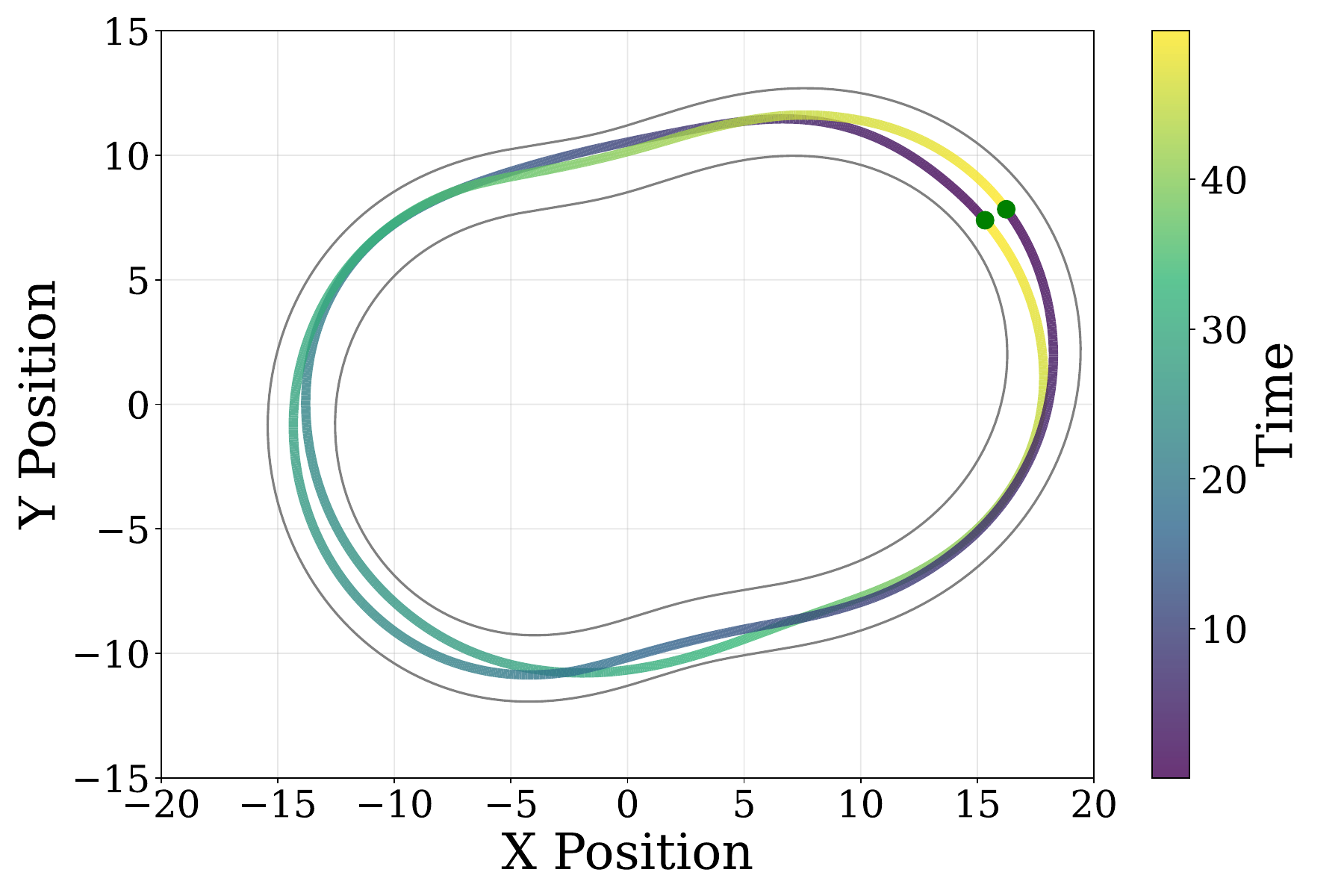}
		\caption{Optimized}
		\label{subfigTrackOptimal}
	\end{subfigure}\hfill\hfill
	\begin{subfigure}{0.475\linewidth}
		\centering
		\includegraphics[width=1\linewidth,height = 0.7\linewidth]{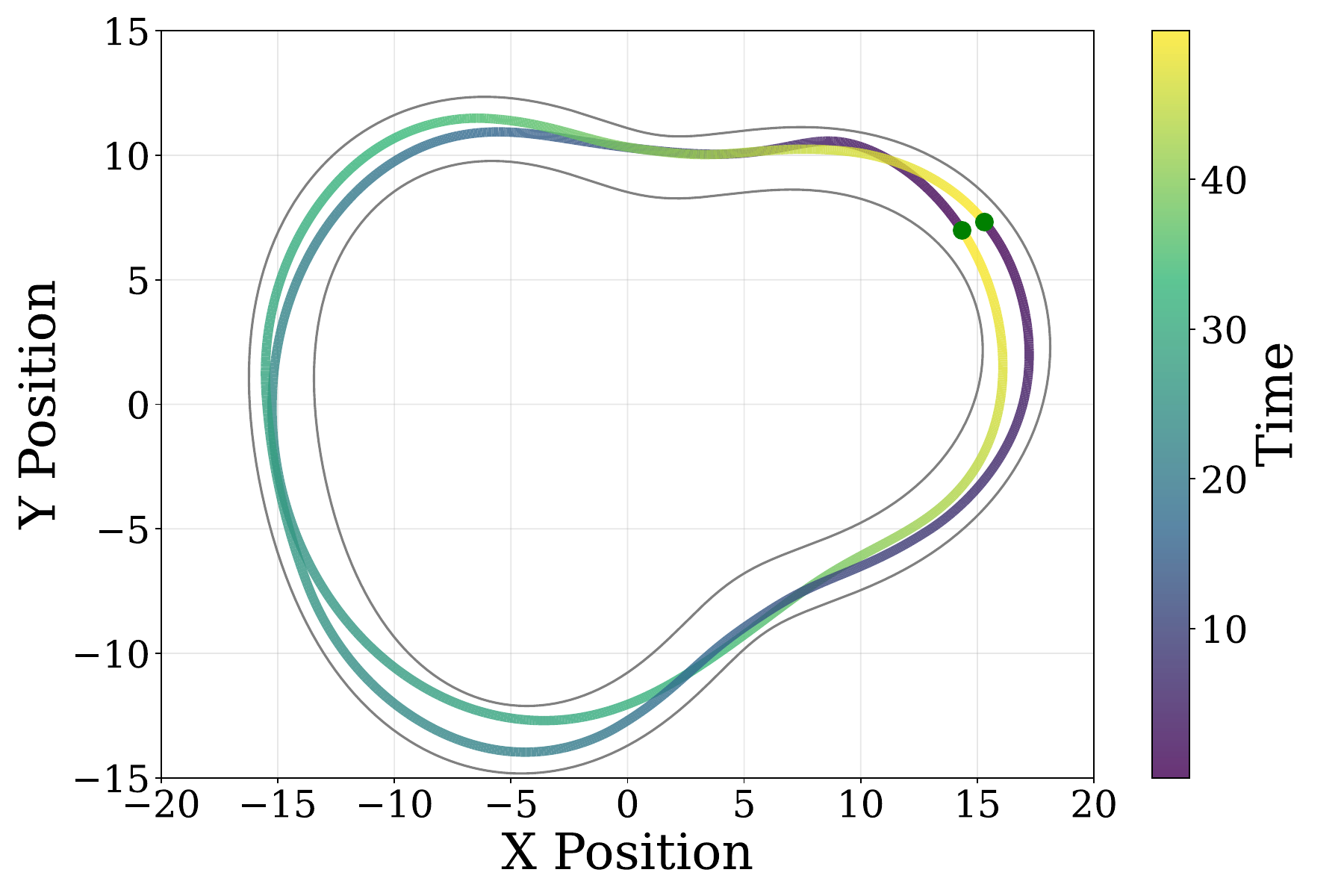}
		\caption{Baseline}
		\label{subfigTrackInitial}
	\end{subfigure}
    \caption{\textbf{(a)} Optimized environment that guides agents to move smoothly along the track. \textbf{(b)} Baseline environment.}\label{fig:TrackDesign}\vspace{-4mm}
\end{figure}

Table \ref{tab:scenarios} compares the performance, and Figs. \ref{subfigTrackOptimal}-\ref{subfigTrackInitial} show the optimized track and the initial track. In this scenario, since agents' starting and goal positions overlap, the minimum traveled distance is cumbersome to compute. We replace SPL with the ratio of traveled distances between the optimized and baseline environments, where a lower value represents higher navigation efficiency. Our method improves the safety level, decreases the traveled distance, and reduces the energy consumption, while, at the same time, requiring less computation time for trajectory optimization. Moreover, the optimized track guides agents to move more smoothly along the track, with fewer abrupt 
changes in moving directions and velocities.

\section{Conclusion}

This paper formulated a bi-level environment-trajectory co-optimization problem, which comprises a lower-level sub-problem of trajectory optimization and an upper-level sub-problem of environment optimization. The former generates agent trajectories that maximize path efficiency and minimize control effort, while the latter builds upon the former to optimize environment configurations that maximize the safety level of multi-agent navigation. We proposed a continuous metric for the upper-level environment optimization, which measures an explicit safety level of the environment w.r.t. multi-agent navigation. By incorporating the proposed safety metric, we developed a differentiable optimization method, which iteratively tackles the lower-level trajectory optimization with interior point methods and the upper-level environment optimization with gradient ascent.  

Due to the challenge of modeling the relationship between the environment and agents, the connection between the upper- and lower-level sub-problems is unclear. To overcome this issue, we leverage KKT conditions and the IFT to compute the gradients of agent trajectories w.r.t. environment parameters, bridging the two sub-problems in a differentiable manner. Results across Scenarios 1-6 demonstrate the applicability of the developed method in various practical applications inspired from warehouse logistics to urban transportation. These experiments corroborate the theoretical findings and show the effectiveness of our method, i.e., co-optimizing environment configurations and agent trajectories facilitates safe multi-agent navigation. Moreover, our results indicate that appropriately designed obstacles can provide ``positive'' guidance for agent de-confliction, rather than merely imposing ``negative'' hindrance as traditionally assumed. 

Future work will consider the following aspects. First, we can consider more advanced gradient-based methods for the upper-level environment optimization, such as Newton's method, quasi-Newton method, and momentum gradient ascent. This allows us to increase convergence rates and further improve performance. Second, we plan to impose constraints on the upper-level environment optimization, such as restrictions on obstacle changes, and incorporate a primal-dual mechanism to handle these constraints. 

\appendices 

\begin{table*}[t]
\scriptsize
\centering
\caption{Performance of the proposed method and baselines with unicycle dynamics.}
\label{tab:unicycle}
\setlength{\tabcolsep}{3pt} 
\begin{tabular}{llccccc}
\toprule
\textbf{Scenario} &  & \textbf{Safety metric $\uparrow$} & \textbf{SPL $\uparrow$} & \textbf{NumCOLL $\downarrow$} & \textbf{Computation time $\downarrow$} & \textbf{PCTSpeed $\downarrow$} \\
\midrule
\multirow{3}{*}{Narrow Passage (unicycle)} 
    & Optimized environment (ours) & \textbf{0.833} & 0.981 & \textbf{0} & \textbf{41.275} & \textbf{0.227} \\
    & Narrow environment & 0.777 & 0.756 & 0.167 & 57.331 & 0.229 \\
    & Wide environment & 0.784 & \textbf{0.987} & \textbf{0} & 74.580 & 0.275 \\
\midrule
\multirow{3}{*}{Road intersection (unicycle)} 
    & Optimized environment (ours) & \textbf{0.949} & \textbf{0.990} & 0 & \textbf{26.043} & \textbf{0.701} \\
    & Baseline environment with an intersection angle $\pi/4$ & 0.932 & 0.984 & 0 & 56.251 & 0.705 \\
    & Baseline environment with an intersection angle $3\pi/4$ & 0.944 & 0.988 & 0 & 29.213 & \textbf{0.701} \\
\bottomrule
\end{tabular}
\end{table*}

\section{Proofs of Safety Metric Properties}\label{appendix:proofs}

We prove the three properties in Section \ref{subsec:properties}.

\noindent \emph{\textbf{(i)}} From the definitions of safety w.r.t. obstacle regions $s_\Delta(\ccalA)$ [cr. \eqref{eq:safetyObstaclesCom}] and safety among agents $s_\ccalA(\ccalA)$ [cf. \eqref{eq:safetyAgentCom}], we have
\begin{align}\label{eq:nonnegative}
    p_\Delta(\ccalA) \ge 0,~ p_\ccalA(\ccalA) \ge 0.
\end{align}
From the definition of the safety metric [cf. \eqref{eq:safetyTuple}] and \eqref{eq:nonnegative}, we complete the proof $\ccalP_\ccalE(\ccalA)\ge 0$.

\emph{\textbf{(ii)}} For any agent $A_i$ in a multi-agent system $\ccalA_k$, we have
\begin{align}
    A_i \in \ccalA_k \in \ccalA,
\end{align}
where $\ccalA = \cup_{k=1}^K \ccalA_k$ is the union of the multi-agent systems $\{\ccalA_k\}_{k=1}^K$. With the same obstacle regions $\{\Delta_j\}_{j=1}^M$, we have
\begin{align}
    \sum_{i \in \ccalA} \sum_{j=1}^M \sum_{t=0}^T p_{\Delta_j}(\bbx_i^{(t)}) = \sum_{k=1}^K \sum_{i \in \ccalA_k} \sum_{j=1}^M \sum_{t=0}^T p_{\Delta_j}\!(\bbx_i^{(t)}).
\end{align}
From the definition of the unsafety mass w.r.t. obstacle regions [cf. \eqref{eq:multiSafetyObstacle}], it holds that
\begin{align}\label{proof:obstacleSafety}
	\Big(\sum_{k=1}^K N_k\Big) M \cdot p_\Delta(\ccalA) = \sum_{k=1}^K \Big(N_k M \cdot p_\Delta(\ccalA_k)\Big). 
\end{align}
For any other agent $A_{i'}$ in a different multi-agent system $\ccalA_{k'}$ with $k' \ne k$, since $\ccalA_{k'}$ and $\ccalA_k$ are disjoint, there is no collision zone between agents $A_{i'}$ and $A_i$, we have $\sum_{t=0}^T p_{\bbx_{i'}}(\bbx_i^{(t)}) = 0$ by definition [cf. \eqref{eq:SafetyAgent1}] and thus, we have 
\begin{align}
    \sum_{i' \ne i \in \ccalA} \sum_{t=0}^T p_{\bbx_{i'}}(\bbx_i^{(t)}) = \sum_{i'' \ne i \in \ccalA_k} \sum_{t=0}^T p_{\bbx_{i''}}(\bbx_i^{(t)}).
\end{align}
From the definition of the unsafety mass among agents [cf. \eqref{eq:multiSafetyAgent}], it holds that
\begin{align}\label{proof:agentSafety}
	\Big(\sum_{k\!=\!1}^K\! N_k\!\Big) \Big( \sum_{k=1}^K\! N_k\!-\!1\! \Big)\! \cdot\! p_\ccalA\big(\ccalA\big) \!=\!\! \sum_{k=1}^K\! \Big(N_k (N_k \!-\! 1) \!\cdot\! p_{\ccalA_k}\!(\ccalA_k)\!\Big). 
\end{align}
By using \eqref{proof:obstacleSafety} and \eqref{proof:agentSafety} in the definition of the safety metric [cf. \eqref{eq:safetyTuple}], we complete the proof
\begin{align}
    \Big(\sum_{k=1}^K N_k\Big) \cdot \ccalP_\ccalE(\ccalA) = \sum_{k=1}^K \Big(N_k \cdot \ccalP_\ccalE(\ccalA_k)\Big). 
\end{align}

\emph{\textbf{(iii)}} For any agent $A_i$ in a multi-agent system $\ccalA_k$, we have
\begin{align}
    A_i \in \ccalA_k \subseteq \ccalA,
\end{align}
where $\ccalA = \cup_{k=1}^K \ccalA_k$ is the union of the multi-agent systems $\{\ccalA_k\}_{k=1}^K$. For any other agent $A_{i'}$ in a different multi-agent system $\ccalA_{k'}$ with $k' \ne k$, we have
\begin{align}
    &\sum_{i' \ne i \in \ccalA} \sum_{t=0}^T p_{\bbx_{i'}}(\bbx_i^{(t)}) \\
    &= \sum_{i' \ne i \in \ccalA_k} \sum_{t=0}^T p_{\bbx_{i'}}(\bbx_i^{(t)}) \!+\!\! \sum_{i' \ne i \in (\ccalA \setminus \ccalA_k)} \sum_{t=0}^T p_{\bbx_{i'}}(\bbx_i^{(t)}). \nonumber
\end{align}
Since $p_{\bbx_{i'}}(\bbx_i^{(t)}) \ge 0$ for any $i, i'$ and $t$, it holds that 
\begin{align}
    &\sum_{i' \ne i \in \ccalA} \sum_{t=0}^T p_{\bbx_{i'}}(\bbx_i^{(t)}) \ge \sum_{i' \ne i \in \ccalA_k} \sum_{t=0}^T p_{\bbx_{i'}}(\bbx_i^{(t)}).
\end{align}
From the definition of the unsafety mass among agents [cf. \eqref{eq:multiSafetyAgent}], it holds that
\begin{align}\label{proof:agentSafety1}
	\Big(\!\sum_{k\!=\!1}^K N_k\!\Big)\! \Big( \sum_{k\!=\!1}^K N_k\!-\!1 \!\Big) \!\cdot\! p_\ccalA\big(\ccalA\big) \!\ge\! \sum_{k\!=\!1}^K\! \Big(\!N_k (N_k \!-\! 1) \!\cdot\! p_{\ccalA_k}\!(\ccalA_k)\!\Big). 
\end{align}
By using \eqref{proof:obstacleSafety} and \eqref{proof:agentSafety1} in the definition of the safety metric [cf. \eqref{eq:safetyTuple}], we complete the proof
\begin{align}
    \sum_{k=1}^K \Big(N_k \cdot \ccalP_\ccalE(\ccalA_k)\Big) \le \Big(\sum_{k=1}^K N_k\Big) \cdot \ccalP_\ccalE(\ccalA). 
\end{align}

\begin{figure}
	\centering
	\begin{subfigure}{0.475\columnwidth}
		\includegraphics[width=1\linewidth, height = 0.7\linewidth]{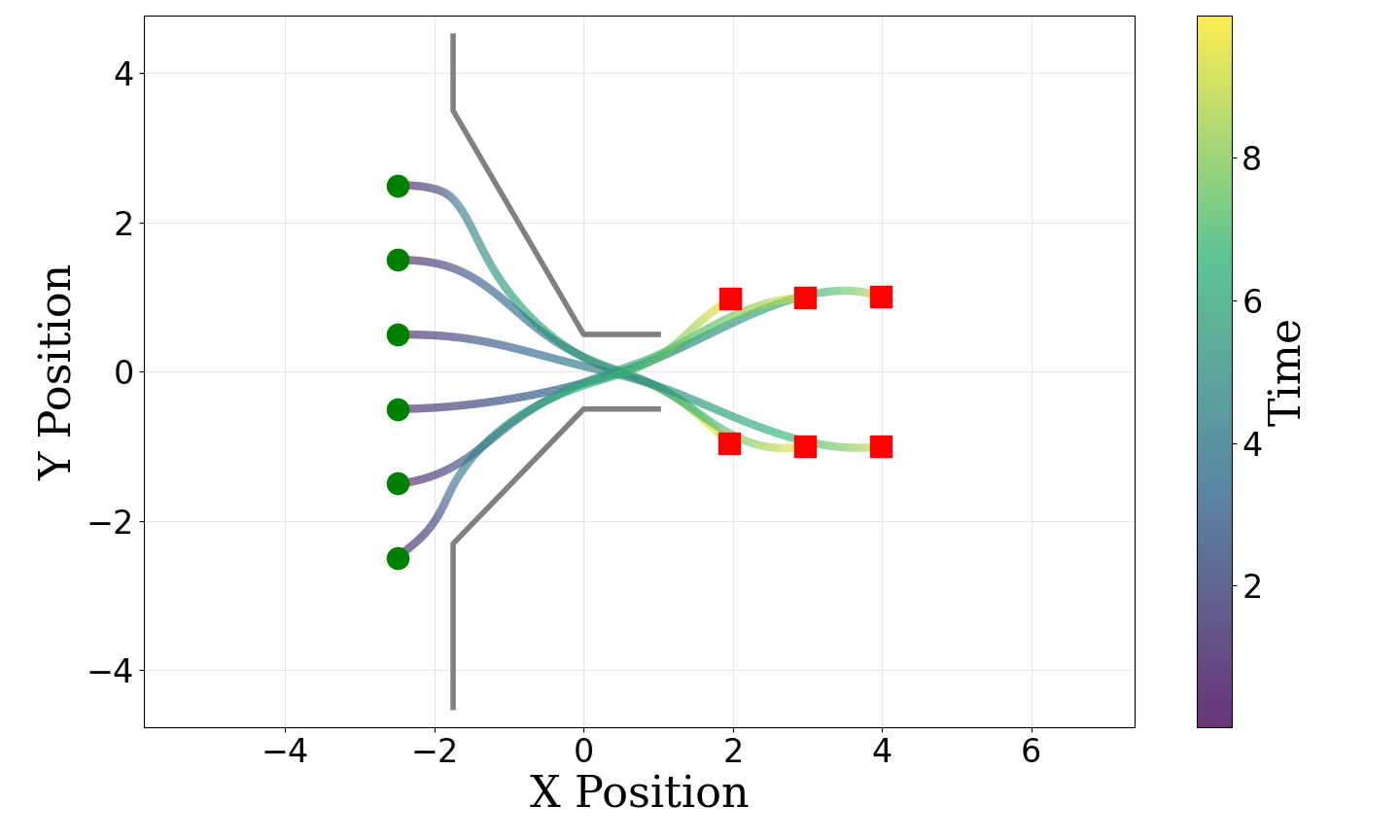}
		\caption{}
		\label{subfigNarrowbCycle}
	\end{subfigure}\hfill\hfill
	\begin{subfigure}{0.475\columnwidth}
		\includegraphics[width=1\linewidth,height = 0.7\linewidth]{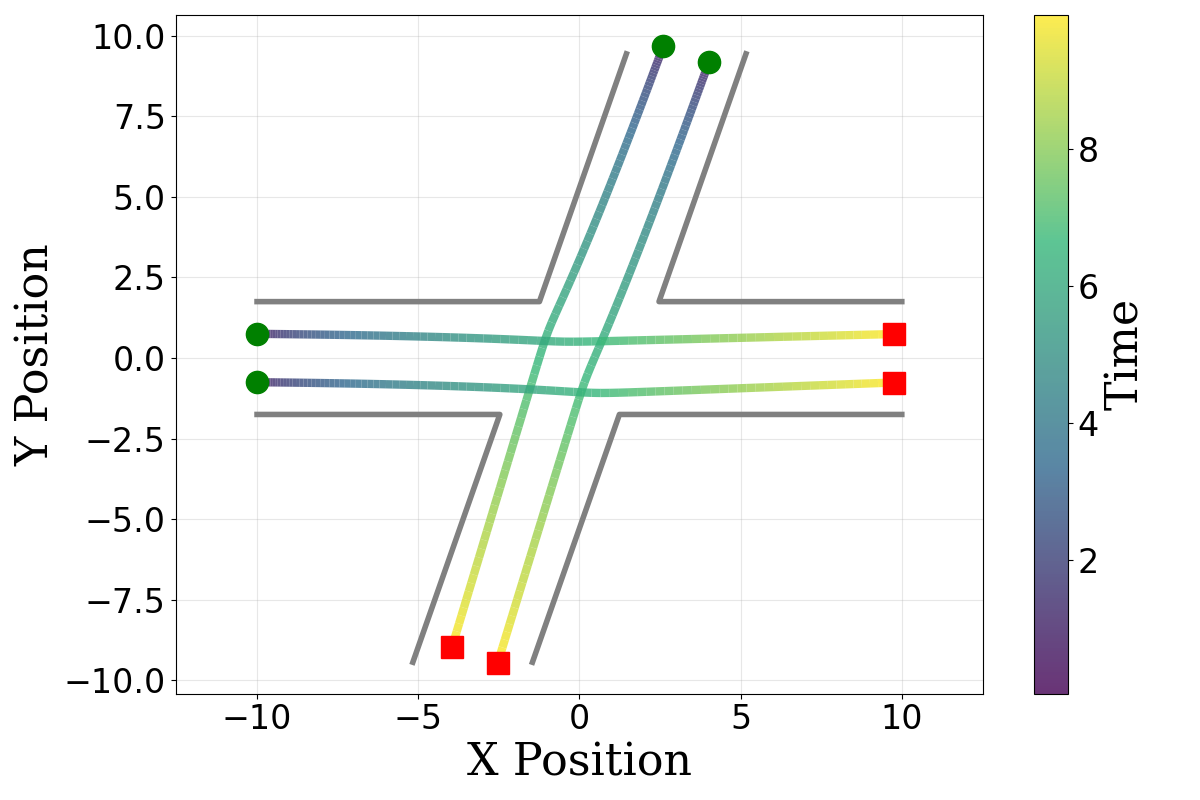}
		\caption{}
		\label{subfigNarrowdCycle}
	\end{subfigure}
	\caption{\textbf{(a)} Optimized environment with unicycle dynamics in the scenario of Narrow Passage. \textbf{(b)} Optimized environment with unicycle dynamics in the scenario of Road Intersection.}\label{fig:Cycle}\vspace{-4mm}
\end{figure} 

\section{Additional Experiments}\label{append:addExp}

We conduct additional experiments to study the role of agent dynamics [cf. \eqref{eq:dynamics}] in environment-trajectory co-optimization and to explore an extension of our bi-level framework to stochastic environment-trajectory co-optimization. 

\begin{table*}[t]
\scriptsize
\centering
\caption{Performance of the proposed method and baselines for stochastic environment-trajectory co-optimization.}
\label{tab:stochasticEnv}
\setlength{\tabcolsep}{3pt} 
\begin{tabular}{lccccc}
\toprule
\textbf{Stochastic environment optimization} & \textbf{Safety metric $\uparrow$} & \textbf{SPL $\uparrow$} & \textbf{NumCOLL $\downarrow$} & \textbf{Computation time $\downarrow$} & \textbf{PCTSpeed $\downarrow$} \\
\midrule
Optimized environment (ours) & \textbf{0.929 $\pm$ 0.013} & \textbf{0.943 $\pm$ 0.019} & \textbf{0 $\pm$ 0} & \textbf{10.465 $\pm$ 9.490} & \textbf{0.816 $\pm$ 0.029} \\
Standard environment with a regular layout & 0.896 $\pm$ 0.029 & 0.916 $\pm$ 0.020 & 0 $\pm$ 0 & 36.572 $\pm$ 42.818 & 0.826 $\pm$ 0.021 \\
Baseline environment with a random layout & 0.894 $\pm$ 0.035 & 0.923 $\pm$ 0.022 & 0.075 $\pm$ 0.225 & 71.270 $\pm$ 86.946 & 0.821 $\pm$ 0.027 \\
\bottomrule
\end{tabular}
\end{table*}

\subsection{System Dynamics}\label{subsec:dynamics}

System dynamics determine how agents' states transit with their actions between successive time steps, and have a significant impact on the lower-level trajectory optimization. Since the upper-level environment optimization relies heavily on agent trajectories, different dynamics may require distinct environment configurations for safe multi-agent navigation. To investigate this impact, we switch from double integrator dynamics to unicycle dynamics, and evaluate our method in narrow passage and road intersection scenarios. 

Table \ref{tab:unicycle} shows the performance comparison and Fig. \ref{fig:Cycle} displays the optimized environments in two scenarios. We see that different dynamics lead to distinct optimal environments and agent trajectories, corroborating the impact of agent dynamics. The optimized environment outperforms the baselines for unicycle dynamics as well, corresponding to that for double integrator dynamics in Section \ref{subsec:performance}. These results demonstrate the general applicability of our method in varying dynamics and emphasize the benefits of differentiable environment-trajectory co-optimization for safe multi-agent navigation. 

\subsection{Stochastic Co-Optimization}

The proposed differentiable bi-level framework can be extended for environment-trajectory co-optimization over random multi-agent navigation tasks, referred to as stochastic environment-trajectory co-optimization. Specifically, for random tasks with initial and goal positions sampled from a distribution $\ccalD$, we can formulate a stochastic bi-level optimization problem as 
\begin{align}\label{eq:Stobilevel}
    \max_{\bbtheta \in \mathbb{O}} \quad &
    \mathbb{E}_{\ccalD}[F(\bbx^\star(\bbtheta), \bbu^\star(\bbtheta), \bbtheta)] \\
    \text{s.t.} \quad & \text{Random task}~~ \{\bbs_i, \bbg_i\}_{i=1}^N \sim \ccalD \nonumber \\
    &
     \text{Obstacle compliance}~~ G\big(\bbx^\star(\bbtheta), \bbu^\star(\bbtheta), \bbtheta\big) \leq 0 \nonumber \\
    &
    \bbx^\star(\bbtheta), \bbu^\star(\bbtheta) = \argmin_{\bbx \in \mathbb{X},\bbu \in \mathbb{U}} f(\bbx, \bbu) \nonumber \\
    & \text{s.t.} ~~ \text{Agent dynamics}~~ \bbx_i^{(t)} = \Phi(\bbx_i^{(t-1)}, \bbu_i^{(t-1)}, dt) \nonumber \\
    & \qquad \text{Obstacle avoidance} 
    ~~ g_{o,i}\big(\bbx_i^{(t)}, \bbtheta\big) \le 0 \nonumber \\
    & \qquad \text{Agent avoidance}~~g_{a}\big(\bbx_i^{(t)}, \bbx_{i'}^{(t)}\big) \le 0~\for~i \neq i' \nonumber \\
    & \qquad \text{Initial conditions} 
    ~~ \bbx_{i}^{(0)} = \bbs_{i}.& \nonumber
\end{align}
The upper-level goal becomes to optimize environment parameters $\bbtheta^\star$ that maximize the expected safety metric over random multi-agent navigation tasks $\mathbb{E}_{\ccalD}[F(\bbx^\star(\bbtheta), \bbu^\star(\bbtheta), \bbtheta)]$. We consider Scenario 1 Warehouse with $4$ agents and $9$ obstacles. There are $19$ possible starting positions distributed along the left $\&$ top boundaries, and $19$ possible goal positions distributed along the bottom $\&$ right boundaries. Two agents are randomly initialized at the left boundary and tasked towards the right boundary, while the other two agents are randomly initialized at the top boundary and tasked towards the bottom boundary, ensuring non-trivial navigation tasks. 

\begin{figure}%
	\centering
	\includegraphics[width=0.525\linewidth,height = 0.35\linewidth]{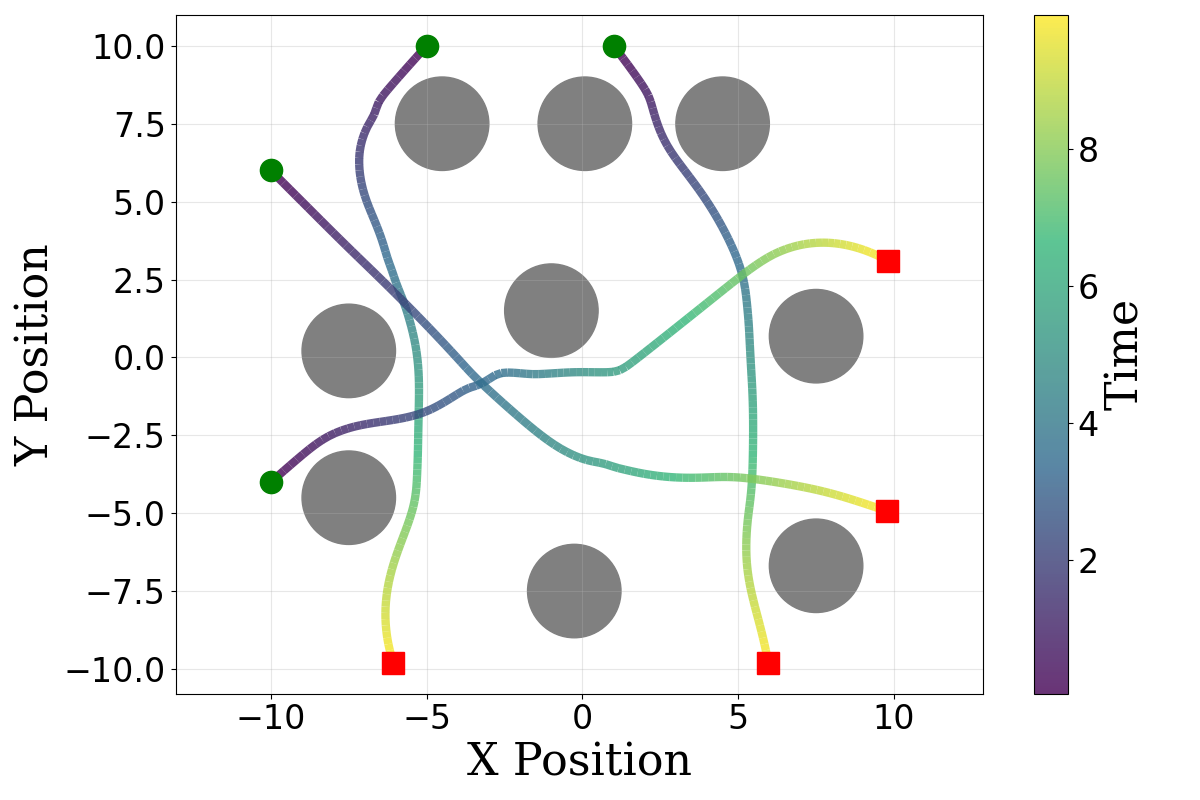}
	\caption{Optimized environment w.r.t. random navigation tasks, where the agent trajectories shown in the figure are generated based on one sampled navigation task.}\label{fig:average}\vspace{-4mm}
\end{figure}

Table \ref{tab:stochasticEnv} shows the results and Fig. \ref{fig:average} displays the optimized environment. Our method outperforms the baselines in the stochastic setting as well. It demonstrates a higher safety level, a lower NumCOLL, and less computation time, which validates the effectiveness of our method over random navigation tasks, and achieves a higher SPL and a lower PCTSpeed, which improves the expected navigation performance in the meantime. Moreover, the optimized environment exhibits an irregular / asymmetric obstacle layout to a certain degree, which differs from the regular / symmetric environment as is common practice. This emphasizes the challenge of hand-designing environments with human intuition and highlights the significance of our differentiable optimization method. 

\renewcommand*{\bibfont}{\footnotesize}
\printbibliography

\end{document}

%% file: mySections.tex
\usepackage{needspace}



%% file: colors.tex
\definecolor{penndarkestblue}{cmyk}{1,0.74,0,0.77}
\definecolor{penndarkerblue}{cmyk}{1,0.74,0,0.70}
\definecolor{pennblue}{cmyk}{0.99,0.66,0,0.57} 
\definecolor{pennlighterblue}{cmyk}{0.98,0.44,0,0.35}
\definecolor{pennlightestblue}{cmyk}{0.38,0.17,0,0.17} 

\definecolor{penndarkestred}{cmyk}{0,1,0.89,0.66}
\definecolor{penndarkerred}{cmyk}{0,1,0.88,0.55}
\definecolor{pennred}{cmyk}{0,1,0.83,0.42} 
\definecolor{pennlighterred}{cmyk}{0,1,0.6,0.24}
\definecolor{pennlightestred}{cmyk}{0,0.43,0.26,0.12} 

\definecolor{penndarkestgreen}{cmyk}{1,0,1,0.68}
\definecolor{penndarkergreen}{cmyk}{1,0,1,0.57}
\definecolor{penngreen}{cmyk}{1,0,1,0.44} 
\definecolor{pennlightergreen}{cmyk}{1,0,1,0.25}
\definecolor{pennlightestgreen}{cmyk}{0.43,0,0.43,0.13}

\definecolor{penndarkestorange}{cmyk}{0,0.65,1,0.49}
\definecolor{penndarkerorange}{cmyk}{0,0.65,1,0.33}
\definecolor{pennorange}{cmyk}{0,0.54,1,0.24} 
\definecolor{pennlighterorange}{cmyk}{0,0.32,1,0.13}
\definecolor{pennlightestorange}{cmyk}{0,0.15,0.46,0.06}

\definecolor{penndarkestpurple}{cmyk}{0,1,0.11,0.86}
\definecolor{penndarkerpurple}{cmyk}{0,1,0.13,0.82}
\definecolor{pennpurple}{cmyk}{0,1,0.11,0.71} 
\definecolor{pennlighterpurple}{cmyk}{0,1,0.05,0.46}
\definecolor{pennlightestpurple}{cmyk}{0,0.35,0.02,0.23}

\definecolor{pennyellow}{cmyk}{0,0.20,1,0.05} 
\definecolor{pennlightgray1}{cmyk}{0,0,0,0.05}
\definecolor{pennlightgray3}{cmyk}{0.01,0.01,0,0.18}
\definecolor{pennmediumgray1}{cmyk}{0.04,0.03,0,0.31}
\definecolor{pennmediumgray4}{cmyk}{0.08,0.06,0,0.54}
\definecolor{penndarkgray2}{cmyk}{0.09,0.07,0,0.71}
\definecolor{penndarkgray4}{cmyk}{0.1,0.1,0,0.92}